\documentclass[10pt,journal,compsoc,pagebackref,breaklinks=true,colorlinks=true,citecolor=green,urlcolor=magenta,linkcolor=blue,bookmarks=false]{IEEEtran}

\ifCLASSOPTIONcompsoc
  \usepackage[nocompress]{cite}
\else
  \usepackage{cite}
\fi
\ifCLASSINFOpdf
\else
\fi

\usepackage{booktabs}
\usepackage{graphicx}
\usepackage{multirow} 
\usepackage{microtype}
\usepackage{tikz}
\usepackage[linesnumbered,lined,boxed,commentsnumbered,ruled]{algorithm2e}
\usepackage[pagebackref,breaklinks=true,colorlinks,citecolor=blue,urlcolor=blue,linkcolor=blue,bookmarks=false]{hyperref}

\usepackage{color} 

\usepackage{ragged2e}  

\usepackage{booktabs}
\usepackage{makecell}

\usepackage[capitalize]{cleveref}
\crefname{section}{Sec.}{Secs.}
\Crefname{section}{Section}{Sections}
\Crefname{table}{Table}{Tables}
\crefname{table}{Tab.}{Tabs.}
\crefname{algorithm}{Algo.}{Algos.}

\usepackage{xspace}
\makeatletter
\DeclareRobustCommand\onedot{\futurelet\@let@token\@onedot}
\def\@onedot{\ifx\@let@token.\else.\null\fi\xspace}

\def\etal{\emph{et al}\onedot}

\usepackage{forest}
\usepackage{tikz}

\definecolor{darkpastelgreen}{rgb}{0.01, 0.75, 0.24}
\definecolor{lightcoral}{rgb}{0.94, 0.5, 0.5}
\definecolor{myblue}{HTML}{BCDBF6}
\definecolor{mygreen}{rgb}{0.678, 0.875, 0.690}
\definecolor{myred}{rgb}{0.949, 0.792, 0.780}
\definecolor{myyellow}{rgb}{0.976, 0.953, 0.600}
\definecolor{mycream}{rgb}{1.000, 0.957, 0.765}

\usepackage{xcolor}     
\usepackage{graphicx}   
\usepackage{hyperref}   

\hypersetup{
    colorlinks = true,        
    citecolor  = blue,        
    linkcolor  = red,         
    urlcolor   = green,       
    filecolor  = magenta      
}

\usepackage{enumitem}
\setlist[itemize]{leftmargin=*}

\begin{document}
\title{From Perception to Cognition: A Survey of Vision-Language Interactive Reasoning in Multimodal Large Language Models}
%
%
%
%
%

\author{%
Chenyue~Zhou,
Mingxuan~Wang,
Yanbiao~Ma\textsuperscript{*},
Chenxu~Wu,
Wanyi~Chen,
Zhe~Qian,
Xinyu~Liu,
Yiwei~Zhang,
Junhao~Wang,
Hengbo~Xu,
Fei~Luo,
Tianyi~Jiang,
Xiaohua~Chen,
Xiaoshuai~Hao,
Hehan~Li,
Andi~Zhang,
Wenxuan~Wang,
Kaiyan~Zhang,
Guoli Jia,
Lingling~Li,~\IEEEmembership{Senior Member,~IEEE},\\
Zhiwu~Lu,~\IEEEmembership{Senior Member,~IEEE},
Yang~Lu\textsuperscript{*},~\IEEEmembership{Senior Member,~IEEE},~and~Yike~Guo,~\IEEEmembership{Fellow,~IEEE}%
\IEEEcompsocitemizethanks{%
  \IEEEcompsocthanksitem Yanbiao Ma is with the Gaoling School of Artificial Intelligence, Renmin University of China.\protect\\
  E-mail: ybma1998@ruc.edu.cn
  \IEEEcompsocthanksitem  Yang Lu is with Xiamen University, China.
  \IEEEcompsocthanksitem Yike Guo is with The Hong Kong University of Science and Technology, Hong Kong SAR, China.
  \IEEEcompsocthanksitem  Chenyue Zhou is with Nanyang Technological University, Singapore.
}
\thanks{Manuscript received September 28, 2025.}
}

%
%

\markboth{{Journal of \LaTeX\ Class Files, September 2025}}
{Shell \MakeLowercase{\textit{et al.}}: Bare Advanced Demo of IEEEtran.cls for IEEE Computer Society Journals}

\IEEEtitleabstractindextext{
\begin{abstract}

Multimodal Large Language Models (MLLMs) strive to achieve a profound, human-like understanding of and interaction with the physical world, but often exhibit a shallow and incoherent integration when acquiring information (Perception) and conducting reasoning (Cognition). This disconnect leads to a spectrum of reasoning failures, with hallucination being the most prominent. Collectively, these issues expose a fundamental challenge: the ability to process pixels does not yet confer the ability to construct a coherent, credible internal world model.
To systematically dissect and address this challenge, this survey introduces a novel and unified analytical framework: ``From Perception to Cognition." We deconstruct the complex process of vision-language interactive understanding into two interdependent layers: Perception, the foundational ability to accurately extract visual information and achieve fine-grained alignment with textual instructions; and Cognition, the higher-order capability for proactive, multi-step, goal-oriented reasoning built upon this perceptual foundation, the core of which is the formation of a dynamic observe-think-verify reasoning loop.
Guided by this framework, this paper systematically analyzes the key bottlenecks of current MLLMs at both layers. It surveys the landscape of cutting-edge methods designed to address these challenges, spanning from techniques that enhance low-level visual representations to those that improve high-level reasoning paradigms. Furthermore, we review critical benchmarks and delineate future research directions. This survey aims to provide the research community with a clear, structured perspective for understanding the intrinsic limitations of current MLLMs and to illuminate the path toward building next-generation models capable of deep reasoning and a genuine understanding of the world.

\end{abstract}

\begin{IEEEkeywords}
Multimodal Large Language Models (MLLMs), Interactive Vision-Language Reasoning, Perception and Cognition
\end{IEEEkeywords}}

\maketitle

\IEEEdisplaynontitleabstractindextext

%
\IEEEpeerreviewmaketitle

\section{Introduction}
\label{sec:introduction}

\begin{figure*}[t]
    \centering
    \includegraphics[width=1.0\linewidth]{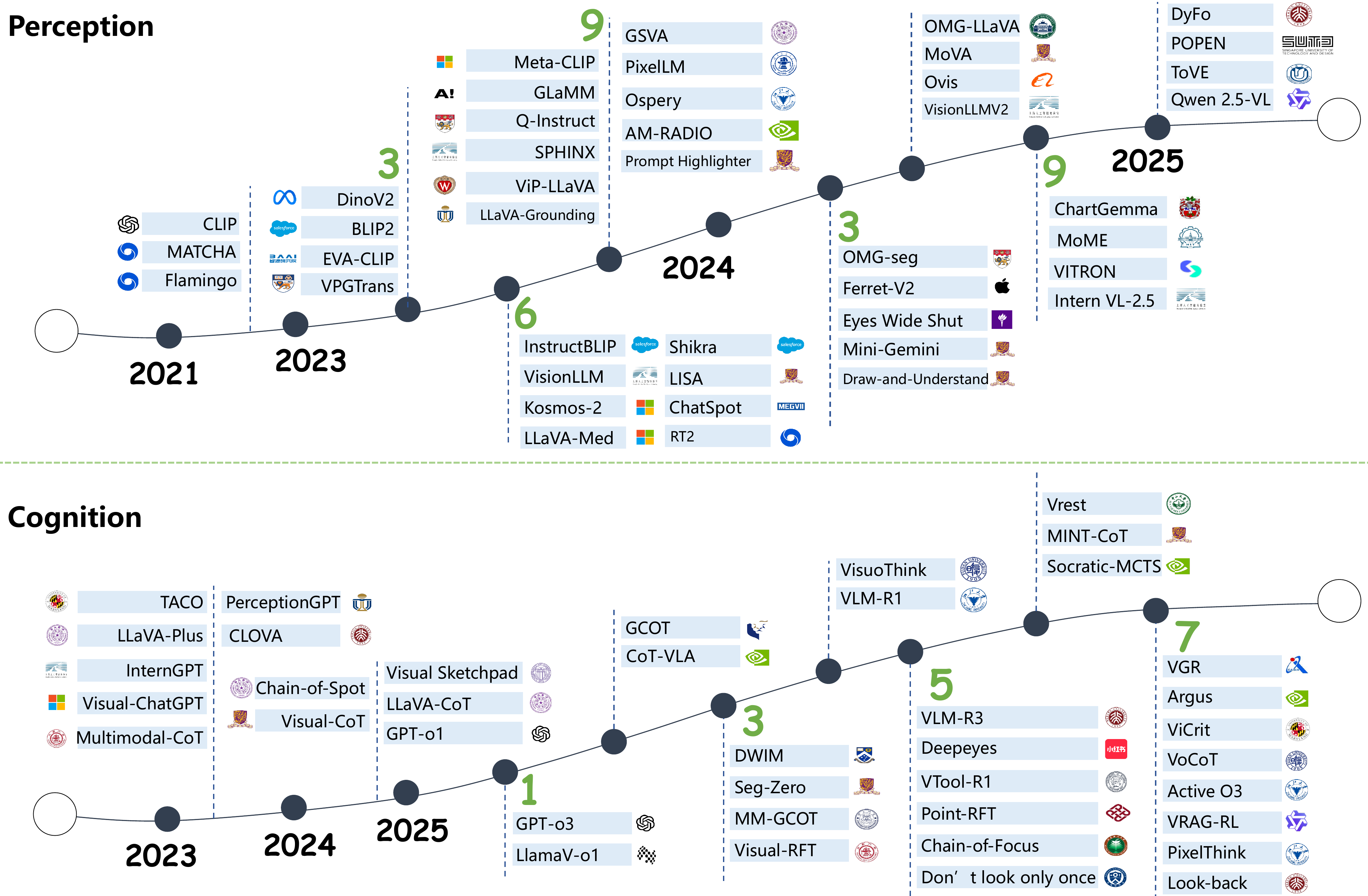}
    \vskip -0.1in
    \caption{Evolution of representative multimodal large language models from 2021 to 2025 organized along Perception and Cognition.}
    \label{fighistory}
\vskip -0.1in
\end{figure*}

\begin{figure*}[t]
    \centering
    \includegraphics[width=1\linewidth]{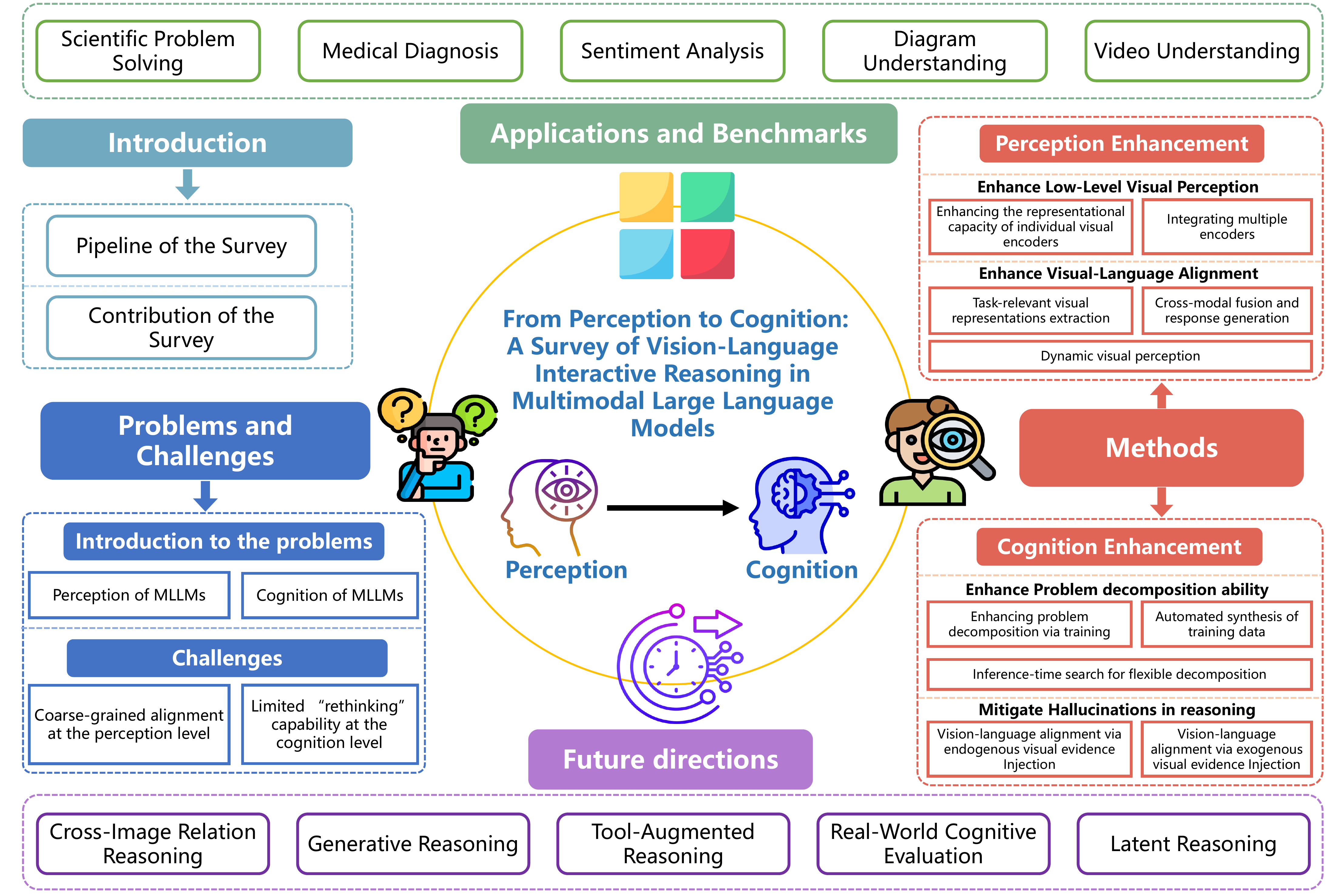}
    \vskip -0.1in
    \caption{The overview of the survey structure.}
    \label{fig:overview}
\vskip -0.1in
\end{figure*}


\IEEEPARstart{T}{he} rapid advancement of Multimodal Large Language Models (MLLMs) is propelling the field of artificial intelligence toward its long-standing goal of Artificial General Intelligence (AGI)~\cite{pei2019towards,bubeck2023sparks,fei2022towards,nmi3}: creating agents that can perceive, reason, and interact with the physical world in a human-like manner~\cite{mni1,mni2,nc1,tayebi2024large}. Central to this progress is the profound synthesis of the sophisticated symbolic reasoning capabilities of Large Language Models (LLMs)~\cite{kasneci2023chatgpt,zhao2023survey} with the potent perceptual acuity of Computer Vision (CV) foundation models~\cite{radford2021learning,simeoni2025dinov3}. On one hand, LLMs such as the GPT series~\cite{brown2020language}, have acquired extensive world knowledge and formidable logical reasoning skills through pre-training on vast text corpora. However, they are inherently confined to a purely symbolic space, operating as ``blind'' reasoners, detached from the sensory richness of the physical world. Conversely, vision foundation models like CLIP~\cite{radford2021learning} have successfully mapped visual and linguistic modalities into a unified embedding space~\cite{zhang2024vision,11050020}, enabling unprecedented perceptual generalization. Yet, they typically lack the deep cognitive faculties required for complex, multi-step reasoning.

The emergence of Multimodal Large Language Models (MLLMs) marked the initial exploration into integrating these two capabilities. In this early Exploration Phase (How to Connect?), exemplified by pioneering works such as Flamingo~\cite{alayrac2022flamingo} and LLaVA~\cite{liu2023visual}, the central challenge was a technical one: how to effectively connect a vision encoder to an LLM. At the time, the research community was primarily focused on solving foundational engineering problems like architectural design and feature alignment, with the primary objective being to make the connection viable.
Once the connection was successfully established, and despite progress on numerous benchmarks, the brittleness of these models became exposed when confronted with scenarios demanding fine-grained perception and complex reasoning. This is specifically manifested in the prevalence of issues such as hallucination and bias in state-of-the-art models~\cite{augenstein2024factuality,ma2024unveiling,nc2,chen2025compositional,xu2025survey}, including Qwen2.5-VL~\cite{bai2025qwen2}, InternVL 2.5~\cite{chen2024expanding}, and GPT-4o~\cite{hurst2024gpt}. These models frequently misinterpret visual details (a perceptual deficit) and are unable to maintain coherent logical chains (a cognitive deficit)~\cite{kalai2025language,11050020}.
Consequently, the research focus has shifted toward a systematic developmental path from Perception to Cognition. In particular, the approach of first establishing a precise \textbf{Perception} layer as a prerequisite for building advanced \textbf{Cognition} has gradually become a consensus for enhancing the capabilities of MLLMs \cite{11050020,tong2024eyes}.

To systematically chart this evolutionary trajectory, this survey introduces the \textbf{``From Perception to Cognition''} framework as a unified lens for analysis. It deconstructs the complex process of vision-language interactive reasoning into two distinct yet interconnected layers. The first, Perception, represents the foundational ability to accurately extract visual information and establish fine-grained alignment with textual instructions. This requires not only recognizing objects, attributes, and relations but also precisely grounding textual concepts to specific visual details. The second, Cognition, is the higher-order capability for proactive, goal-oriented, multi-step reasoning built upon this perceptual foundation. It involves decomposing complex problems, planning logical steps, and, critically, the ability to dynamically re-examine visual evidence to validate or refine its reasoning path, thus forming an \textbf{``observe-think-verify''} feedback loop. As illustrated in Fig.~\ref{fighistory}, we present the timeline of MLLM development from 2021 to 2025, organized along these two central pillars of Perception and Cognition.

Leveraging this framework, this survey provides a structured and comprehensive review of the key problems, methodologies, benchmarks, and future directions in interactive vision-language reasoning. Our review begins by analyzing the core challenges at the perceptual and cognitive levels that have driven the field's progress. Subsequently, we survey the cutting-edge methods aimed at \textbf{enhancing Perception} (e.g., advanced visual encoders) and those designed to \textbf{bolster Cognition} (e.g., sophisticated Chain-of-Thought paradigms and dynamic reasoning mechanisms). By adopting this ``perception-to-cognition” perspective, we aim to elucidate the limitations of existing models and chart a developmental path for technological evolution.

\subsection{Pipeline of the Survey}
This survey presents a comprehensive overview of the key problems, methodologies, benchmarks, and future directions in interactive vision-language reasoning within Multimodal Large Language Models (MLLMs). As illustrated in Fig.~\ref{fig:overview}, our discussion is structured as follows:

\textbf{\cref{sec:Problems and challenges}} outlines the fundamental issues in vision-language interaction. We first establish our core analytical framework by defining \textit{perception} (\cref{subsubsec:Perception}) and \textit{cognition} (\cref{subsubsec:Cognition}). We then use this framework to analyze the primary challenges that MLLMs face along these two dimensions (\cref{subsec:Challenges}).

\textbf{\cref{sec:Methods}} reviews the evolution of relevant methods, organized according to the challenges they address within our perception-cognition framework. 
To solve \textbf{perception-level} problems, we explore techniques focused on enhancing fine-grained visual capabilities (\cref{subsec:how-to-enhance-low-level}) and improving vision-language alignment (\cref{subsec:how-to-enhance-align}).
To address \textbf{cognitive-level} challenges, we delve into methods for enhancing problem decomposition (\cref{subsec:how-to-enhance-decompos}) and mitigating hallucinations by enabling dynamic reasoning that overcomes the limitations of static perceptual memory (\cref{subsec:how-to-mitigate-hallucinations}).
    
\textbf{\cref{sec:Benchmark}} provides a thorough analysis of key benchmarks and applications across diverse domains, including scientific problem-solving, medical diagnosis, diagram understanding, and video reasoning. This section evaluates how current models perform on tasks that require different balances of perceptual and cognitive skills.

\textbf{\cref{sec:Future direction}} concludes the survey by discussing promising future research avenues. We explore emerging paradigms such as latent space reasoning, generative reasoning, and tool-augmented reasoning, highlighting how they might further bridge the gap between perception and cognition.

\subsection{Contribution of the Survey}
Recent years have witnessed a surge in survey papers on Multimodal Large Language Models (MLLMs), each providing a unique perspective on this rapidly evolving field. Among the first, Yin \etal~\cite{yin2024survey} offered a foundational overview of MLLM development up to early 2024.
Subsequent work began to specialize. Some surveys delved into the crucial area of reasoning, analyzing methods for enhancing step-by-step ``slow thinking" processes~\cite{li2025system,wang2025multimodal}. More recently, a number of surveys~\cite{li2025perception,ke2025explain,su2025thinking} has converged on the theme of ``thinking with images," systematically analyzing advancements in fine-grained visual reasoning.

While these surveys provide invaluable insights, they tend to focus either on general MLLM architectures or on specific facets of high-level reasoning. Our survey is distinct from prior work by adopting a more foundational perspective that deconstructs interactive reasoning into two core, interconnected components: perception and cognition. Based on this approach, we make the following key contributions:

\begin{itemize}
    \item \textbf{A Novel Analytical Framework:} We introduce a ``Perception-Cognition" framework that provides a structured lens to understand the fundamental challenges in vision-language interaction. This framework moves beyond a surface-level categorization of tasks or models to dissect the root causes of model failures, such as hallucination, by mapping them to specific deficiencies at either the perceptual or cognitive level.

    \item \textbf{A Structured Taxonomy of Methodologies:} Based on this framework, we provide a systematic and coherent taxonomy of existing methods. We demonstrate how seemingly disparate research efforts (such as enhancing visual encoders and developing advanced Chain-of-Thought) are, in fact, targeted efforts to solve distinct problems along the perception-to-cognition continuum. This clarifies the relationships between different lines of research.
    
\item \textbf{A Unified Perspective on the MLLM Developmental Path:} By structuring our analysis along the Perception-to-Cognition axis, this survey explicitly highlights the fundamental dependency of high-level reasoning on the quality of low-level visual representation. It reframes these two domains not as isolated research areas, but as integral and sequential stages in the development of MLLMs. This unified view offers a holistic roadmap, illustrating that advancements in Perception are a necessary foundation for achieving breakthroughs in Cognition.
\end{itemize}

\section{Problems and challenges}
\label{sec:Problems and challenges}

\begin{figure}[t]
    \centering
    \includegraphics[width=1.0\linewidth]{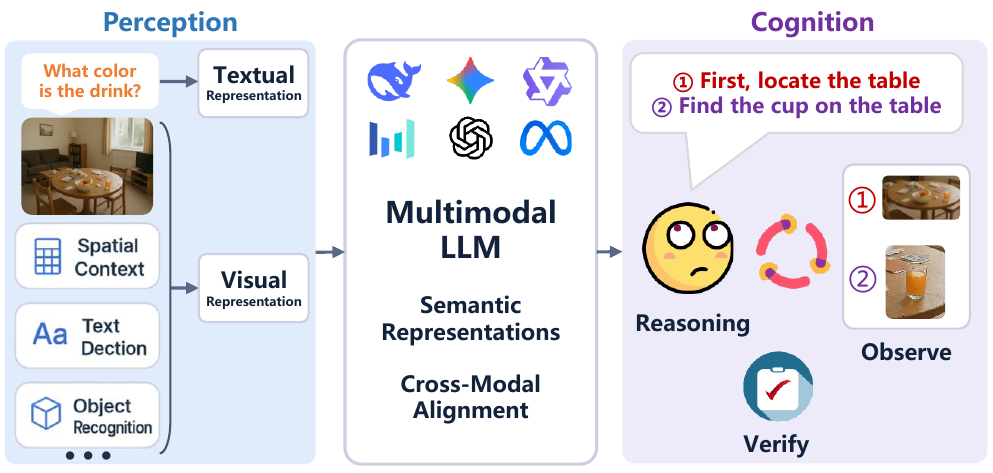}
    \vskip -0.15in
    \caption{Overview of the perception–cognition loop. Perceptual modules extract semantic and spatial evidence which are aligned by a multimodal LLM. Cognition then executes a plan–observe–reason cycle with iterative verification to ground each step in visual evidence.}
    \label{fig:dynamic_relation}
\vskip -0.1in
\end{figure}

Before we deeply explore the classification of methods for vision-language interactive Reasoning in Multimodal large language models (MLLMs), we need to define a core concept to clarify the scope of a model's capabilities in vision-language interactive understanding: \textbf{\emph{What are the perception and cognition of MLLMs ?}} As depicted in Fig.~\ref{fig:dynamic_relation}, we summarize the relationship between perception and cognition.

\subsection{Perception of MLLMs}
\label{subsubsec:Perception}
Perception primarily refers to the ability of multimodal large models to accurately extract relevant visual information from input images when tackling specific visual tasks, and to further encode this raw visual data into semantically meaningful visual representations \cite{he2016deep}, \cite{dosovitskiy2020image}, \cite{radford2021learning}. This process involves not only recognizing visual features such as objects, backgrounds, and texts \cite{yang2021tap}, \cite{xia2023st}, \cite{sidorov2020textcaps} within images, but also discerning spatial relationships between objects, contextual associations \cite{krishna2017visual}, \cite{carion2020end} and deep mining of latent semantic information \cite{kuang2025natural}. More critically, 
effective perception demands that models align these encoded visual representations with input textual information, such as question descriptions, instruction prompts, or dialogue content. This ensures semantic correspondence and mutual reinforcement between visual and textual data. Through such alignment mechanisms, models can provide clear, reliable visual evidence during subsequent vision-language interactions—validating reasoning accuracy or serving as direct answer substantiation. \textbf{In short, perception constitutes the foundational capability for models to ``see" and ``understand" visual information, laying the groundwork for subsequent higher-level cognitive operations.}

\subsection{Cognition of MLLMs}
\label{subsubsec:Cognition}
Compared with perception, which emphasizes information extraction and alignment, cognition focuses more on a model's proactive decision-making and reasoning capabilities driven by visual tasks. Specifically, cognition manifests as a model's ability to systematically determine \emph{\textbf{when to examine visual information}} and \textit{\textbf{which specific image regions to focus on}} based on current task requirements. This selective attention is not random behavior but rather decisions made by the model by integrating existing reasoning processes with visual textual evidence.
After acquiring relevant visual information, the model integrates existing textual and visual data to perform further reasoning. Crucially, cognition also requires the model to dynamically assess the sufficiency of current information at each reasoning step: if existing evidence proves insufficient to support conclusions, the model proactively supplements additional visual evidence (e.g., refocusing on relevant image regions to capture finer features); conversely, if current information adequately supports the inference, it directly generates the final answer. This process forms a closed-loop cycle of ``observation” (acquiring visual information), ``thinking” (multimodal reasoning), and ``verification” (validating the match between reasoning results and evidence). This ensures the model's cognitive process not only maintains logical consistency but also dynamically adjusts its review strategy based on task difficulty, enabling handling of complex vision-language interaction tasks.  

\subsection{Challenges}
\label{subsec:Challenges}

Building on a clear understanding and distinction of perception and cognition in MLLMs, this section will, within this framework, elaborate on the challenges that MLLMs face in vision-language reasoning.
Despite significant advancements in MLLMs for vision-language understanding, where they achieve near-human or even superior performance in tasks like image captioning and visual question answering—a series of critical issues remain unresolved in practical applications. These challenges impose heightened demands on both perception and cognition. \textbf{We group these challenges into two primary categories, corresponding to the domains of perception (\cref{subsubsec:Coarse-grained-vision-language}) and cognition (\cref{subsubsec:Limited-rethinking}).}

\subsubsection{Barriers to Accurate Perception: Weak Extraction and Coarse Alignment}
\label{subsubsec:Coarse-grained-vision-language}

\noindent$\bullet$ \textbf{\emph{Weak Low-Level Visual Information Extraction Capability:}} The fundamental limitation of early mainstream MLLM models, such as the LLaVA~\cite{liu2023visual} \cite{liu2024improved} series, stems from the shortcomings of their underlying CLIP-ViT visual encoders. CLIP-ViT's pretraining objective focuses on global vision-language alignment, learning matches between the visual semantics of entire images and entire text rather than precise regional or pixel-level alignment. This alignment approach leads to poor fine-grained information recognition and weak spatial localization capabilities. Beyond perceptual limitations, CLIP~\cite{radford2021learning} also suffers from specific data generalization issues. The absence of structured, symbolic visual information (e.g., mathematical geometry problems, charts) in its pretraining dataset causes suboptimal performance in scenarios requiring symbolic or structured understanding.

\noindent$\bullet$ \textbf{\emph{Limited Interaction between Visual and Textual Information:}}  Most existing MLLMs first encode images independently before projecting them uniformly into the decoder. Cross-modal interaction primarily relies on global relevance alignment, failing to map language queries to specific regions or pixels. Consequently, the coupled information between text and vision is underutilized.

\subsubsection{Limited Rethinking Capability Constrains Reasoning}
\label{subsubsec:Limited-rethinking}

\noindent$\bullet$ \textbf{\emph{Challenges in Decomposing Complex Vision-Language Interaction Tasks:}} Lacking executable task decomposition paradigms, existing methods predominantly employ single-step or fixed-template approaches, supervising only outcome correctness during training rather than process correctness and consistency.

\noindent$\bullet$ \textbf{\emph{One-Shot Evidence-Based Memory-Based Reasoning Leads to Forgetting and Hallucinations:}} A fundamental limitation of most MLLMs is their static, single-pass approach to visual processing during inference. They perform a single encoding of the image at the input stage and do not revisit it during the subsequent reasoning and text generation process. This lack of a dynamic perception-reasoning loop, where visual evidence can be re-examined as needed, often causes a decoupling between the final answer and the underlying visual facts. This issue is particularly exacerbated in complex, long-chain tasks that require understanding the spatial and semantic relationships among multiple visual elements. Consequently, enabling more robust reasoning requires a dynamic mechanism capable of strategically deciding \textbf{when} to re-examine the image for evidence, \textbf{where} to focus attention, and \textbf{what} specific information to extract at each step of the reasoning process.

In summary, by systematically deconstructing the complex challenges of MLLMs into bottlenecks at the \textbf{perceptual level} and reasoning limitations at the \textbf{cognitive level}, we not only gain a clearer diagnosis of the root causes behind model failures like hallucination but also reveal the underlying logic of the field's technological evolution. It is precisely these specific challenges that constitute the primary driving force behind advancements in MLLMs, spurring a vast body of work aimed at either bolstering perception or deepening cognition.
In the following section, guided by this framework, we will systematically review and analyze the cutting-edge methodologies that the research community has proposed to overcome these distinct perceptual and cognitive challenges.

\section{Methods: a survey}
\label{sec:Methods}
\begin{table*}[t!]
\setlength\tabcolsep{8pt}
\footnotesize
\caption{Representative vision backbone models and their enhanced low-level visual representations.}
\vspace{-5pt}
\centering
\begin{tabular}{l|l|l|p{9.2cm}}
\toprule
Research Direction & Method & Venue  & Main Contribution \\
\midrule
\multirow{3}{*}{\shortstack{Single-encoder \\ enhancement}}
 & \multicolumn{3}{l}{\textit{\textbf{a. Single-encoder optimization.}}} \\
 & EVA-CLIP~\cite{sun2023eva}  & arXiv'23  &   proposing training methods to enhance the efficiency and stability of CLIP in large-scale settings \\
 & SigLip~\cite{zhai2023sigmoid}   & CVPR'23    & Leveraging a sigmoid loss for  enhancing fine-grained representation. \\
 & MetaCLIP~\cite{xu2023demystifying} & ICLR'24      &  Enhancing fine-grained representation by training in larger-scale, high-quality datasets. \\
 & DINOv2~\cite{oquab2023dinov2}   & TMLR'25      & Strengthening geometric and structural representation through self-supervised learning. \\
 & DIVA~\cite{wang2024diffusion}  & ICLR'25  & By conditioning the diffusion model on dense visual features from CLIP and applying reconstruction loss to optimize CLIP.  \\
 & VLV~\cite{zhang2025vision}  & arXiv'25  &Improving CLIP through a unified architecture for image understanding and generation. \\
\midrule
\multirow{6}{*}{\shortstack{Multi-encoder \\Integration  }}

 & \multicolumn{3}{l}{\textit{\textbf{b. Static fusion of encoders.}}} \\
 & Eyes Wide Shut~\cite{tong2024eyes} & CVPR'24    &  Directly concatenating the visual tokens from CLIP and DINOv2 in an alternating, interleaved manner to form a longer sequence. \\
 &Prismatic VLMs~\cite{karamcheti2024prismatic} & ICML'24  & Concatenating the visual tokens from CLIP and DINOv2 along the channel dimension without increasing the sequence length.     \\
 & Ferret-v2~\cite{zhang2024ferret}   & CoLM'24   & Employing multi-granularity fusion of visual representations for fine-grained perception.   \\
 & MouSi~\cite{fanpoly}   & CoLM'24   & Integrating multi-scale encoder outputs, combining global context with local structural cues for robust spatial reasoning.   \\ 
 & BRAVE~\cite{kar2024brave}         & ECCV'24   &  Statically fusing visual features by adopting resolution-aware projection to preserve both high-resolution local detail and low-resolution global semantics.  \\
 & SPHINX~\cite{lin2023sphinx}         & ECCV'24   & Leveraging hierarchical fusion of CLIP and DINO to balance semantic alignment and geometric precision.    \\
 & ParGo~\cite{wang2025pargo}         & AAAI'25   & Introducing local and global tokens with dedicated attention masks to extract both fine-grained and holistic visual information.    \\
 & Layer\_Select\_Fuse~\cite{lin2025multi} & CVPR'25 & Directly fusing multi-layer visual features at the input stage proves to be more stable and effective. \\
\cmidrule{2-4}
 & \multicolumn{3}{l}{\textit{\textbf{c. MoE-based multi-encoder fusion.}}} \\
 & MoME~\cite{shen2024mome}           & NeurIPS'24 & Task-adaptively fusing CLIP and DINOv2 features via a Mixture-of-Experts.  \\
 & MoVA~\cite{zong2024mova}           & NeurIPS'24    & Enhancing cross-task representation flexibility using a multi-expert MoE. \\
 & VisionWeaver~\cite{wang2025diving}      & EMNLP'25  & Dynamically combining SAM, Vary (specialized in text recognition) and DINOv2 through a routing module.    \\
 & TOVE~\cite{wu2025tove}           & ICLR'25    & Feature combination is achieved through coarse-grained context-aware expert routing and a fine-grained expert fusion module. \\
 & R2-T2~\cite{li2025r2}          & ICML'25    & Locally optimizing the routing weights at test time by steering them toward those of correctly predicted samples in the neighborhood. \\
\cmidrule{2-4}
 & \multicolumn{3}{l}{\textit{\textbf{d. Distillation-based models.}}} \\
 & Radio~\cite{ranzinger2024radio}        & CVPR'24   & Enabling a single encoder to absorb heterogeneous vision encoders capabilities via multi-teacher feature distillation.   \\
 & UNIC~\cite{sariyildiz2024unic}        & ECCV'24   & Enabling a universal classifier to inherit cross-task skills from multiple teachers encoders via multi-teacher, layer-wise distillation.  \\
 & MoVE-KD~\cite{cao2025move}        & CVPR'25  &  Enabling a single encoder to inherit the distinct abilities of multiple experts via distillation. \\
 & DUNE~\cite{sariyildiz2025dune}           & CVPR'25   & Compressing multi-expert knowledge into an efficient and powerful lightweight model. \\
\bottomrule
\end{tabular}
\label{tab:vision-backbones}
\vspace{-10pt}
\end{table*}

In the previous chapter, we systematically analyzed the challenges in MLLMs' vision-language interaction understanding based on the dimensions of perception and cognition. This chapter will dive into research approaches based on the issues raised in the previous \cref{subsec:how-to-enhance-low-level} and \cref{subsec:how-to-enhance-align} focusing on addressing perception-related problems, corresponding to the subproblems proposed in \cref{subsubsec:Coarse-grained-vision-language}. \cref{subsec:how-to-enhance-decompos} and \cref{subsec:how-to-mitigate-hallucinations} concentrate on solving cognition-related problems, corresponding to the subproblems proposed in \cref{subsubsec:Limited-rethinking}.

\subsection{Enhance the Low-Level Visual Perception of MLLMs}
\label{subsec:how-to-enhance-low-level}
To address the common limitation of weak capacity in low-level visual information extraction, a large number of recent works have concentrated on advancing visual encoders themselves, with the goal of fundamentally improving the representation of fine-grained visual details. We summarize the related methods in Table \ref{tab:vision-backbones}. This line of progress has evolved along two main directions: \textbf{enhancing the representational capacity of individual visual encoders} and \textbf{integrating multiple encoders in a complementary manner}.

\noindent$\bullet$ \textit{\textbf{Enhance the Representational Capacity of Individual Vision Encoders.}} To address the insufficient representation of detail in early encoders like CLIP ~\cite{radford2021learning}, subsequent research on visual foundation models has focused on two main areas: improving \textbf{fine-grained representation} and \textbf{geometric-texture representation}. 
\textbf{On one hand}, to enhance fine-grained representation, subsequent visual encoders such as MetaCLIP~\cite{xu2023demystifying}, SigLip~\cite{zhai2023sigmoid}, and Eva-CLIP~\cite{sun2023eva} improved semantic alignment and fine-grained recognition abilities by optimizing training objectives and constructing high-quality datasets. \textbf{On the other hand}, to improve geometric-texture representation, several works like DINO series ~\cite{zhang2022dino},~\cite{oquab2023dinov2},~\cite{simeoni2025dinov3} proposed a self-supervised training approach to explore the intrinsic properties of visual data. This enabled them to generate visual representations rich in pixel-level geometric information and texture details, leading to strong performance on tasks requiring sophisticated structural awareness, such as segmentation, localization, and depth estimation.
Furthermore, DIVA~\cite{wang2024diffusion} and VLV~\cite{zhang2025vision} unify image generation and understanding by distilling the superior fine-grained representations from generative models into the CLIP-based visual encoder, thereby bridging the gap between generative and discriminative visual understanding.

\noindent$\bullet$ \textit{\textbf{Multi-Encoder Integration and Distillation.}} 
Given that foundational encoders like CLIP~\cite{radford2021learning} excel at high-level semantic representation but lack fine-grained geometric detail 
, and newer models like DINOv2~\cite{oquab2023dinov2} provide rich structural and pixel-level representations, research naturally progressed towards combining their complementary strengths. Early explorations like Eyes Wide Out~\cite{tong2024eyes}, Prismatic VLMs~\cite{karamcheti2024prismatic}, FerretV2~\cite{zhang2024ferret}, SPHINX~\cite{lin2023sphinx}, MouSi~\cite{fanpoly}, BRAVE~\cite{kar2024brave} proposed a static fusion of features from both CLIP and DINOv2. LLaVA-HR~\cite{luo2024feast} introduces a hybrid-resolution adapter to inject high-resolution features into a low-resolution visual encoder. Mini-Gemini~\cite{li2024mini} employs CLIP-generated tokens as low-resolution queries, which cross-attend to features from a separate high-resolution encoder within localized windows at corresponding spatial positions. ParGo~\cite{wang2025pargo} maps the visual features into Partial tokens, which interact only with a subset of visual features, and Global tokens, which interact with all visual features, thereby enabling effective capture of both local and global information in the image. Recent studies~\cite{lin2025multi} have shown through extensive experiments that directly fusing multi-layer visual features at the input stage achieves more stable and effective performance.

Such fusion mechanisms improve model performance on fine-grained recognition, small object perception, and precise spatial localization—tasks that demand rich and multi-scale visual understanding. However, these approaches rely on static feature integration, which simply stitches together visual representations from different ``expert" encoders without adaptive modulation. This rigid combination may lead to conflicting feature requirements across diverse tasks, limiting the model’s flexibility and contextual adaptability in complex vision-language scenarios.

To address this, more advanced methods such as MoME~\cite{shen2024mome}, MoVA~\cite{zong2024mova}, VisionWeaver~\cite{wang2025diving}, TOVE~\cite{wu2025tove}, R2-T2~\cite{li2025r2} introduced Mixture-of-Experts (MoE)~\cite{chen2022towards} architectures. These models dynamically weight and combine fine-grained and geometric features from different experts based on task requirements, mitigating feature conflicts. While effective, using multiple encoders drastically increases computational costs. As a solution, works such as MoVE-KD~\cite{cao2025move}, Radio~\cite{ranzinger2024radio}, UNIC~\cite{sariyildiz2024unic} and Dune~\cite{sariyildiz2025dune} have employed knowledge distillation. This approach transfers the combined strengths of multiple expert ``teachers" into a single, efficient ``student" encoder. The resulting student model can generate a unified representation that effectively captures both fine-grained details and geometric structure, achieving performance comparable to multi-encoder models but at a significantly lower computational cost.

\begin{table*}[t!]
\setlength\tabcolsep{2pt}
\footnotesize
\caption{Representative methods grouped by research directions. We keep ultra-brief contributions for compactness.}
\renewcommand\arraystretch{1}
\vspace{-5pt}
\centering
\begin{tabular}{l|l|l|p{9.2cm}}
\toprule
Research Direction & Method & Venue & Main Contribution \\
\midrule
\multirow{11}{*}{\shortstack{Task-Relevant Visual\\Representations Extraction}}
 & \multicolumn{3}{l}{\textit{\textbf{(a) Improve the projection layer.}}} \\

  & Honeybee~\cite{cha2024honeybee}             & CVPR'24       & Proposing a flexible and locality-preserving visual projector that balances efficiency with spatial understanding. \\
 & Uni\textendash Med~\cite{zhu2024uni} & NeurIPS’24        & Proposing a CMoE module to manage inter-task relationships via dynamic resource allocation. \\
 & ChartMoE~\cite{xu2024chartmoe}         & ICLR'25  & Proposing a MOE projector for aligning textual and graphical elements in different charts.\\
 & LLaVA\textendash ST~\cite{li2025llava} & CVPR'25 & Proposing a post-processing module that compresses projected video embeddings while preserving their spatio-temporal relationships. \\
 & LLaVA\textendash Octopus~\cite{zhao2025llava} & arXiv'25 & Proposing an instruction-driven adaptive Projector.\\
 & Ovis2.5~\cite{lu2025ovis2}  & arXiv'25   & Proposing a visual embedding table to replace the MLP projection network.         \\
 \cmidrule{2-4} 
 & \multicolumn{3}{l}{\textit{\textbf{(b) Task-specific fine-tuning.}}} \\

 & MATCHA ~\cite{liu2022matcha}          & ACL'23         & Math\textendash centric SFT \\
  & LLaVA\textendash Med~\cite{li2023llava}   &  NeurIPS'23  & Medical SFT \\
 & Q\textendash Instruct~\cite{wu2024q} & CVPR'24    & Low-level perception SFT \\
 & ChartInstruct~\cite{masry2024chartinstruct}      & ACL'24         & Chart SFT \\

\cmidrule{2-4}
 & \multicolumn{3}{l}{\textit{\textbf{(c) Prompt-tuning.}}} \\
 & VPT~\cite{jia2022visual}              & ECCV'22        & Proposing visual prompts for adapting downstream visual tasks \\
 & VPGTrans~\cite{zhang2023vpgtrans}          & NeurIPS'23     & Proposing transferable visual prompts generator across MLLMs \\
 & TVP~\cite{zhang2024exploring}              & CVPR'24        & Proposing joint learnable visual and text prompts for coordinated task adaptation \\
\midrule
\multirow{9}{*}{\shortstack{Cross-Modal Fusion\\and Response Generation}}
 & \multicolumn{3}{l}{\textit{\textbf{(d) Improve instruction encoding paradigm.}}} \\
 & Shikra~\cite{chen2023shikra}           & arXiv'23       & Encoding coordinates in text instruction.\\
 & LLaVA\textendash Grounding~\cite{zhang2024llava} & ECCV'24 & Encoding boxes and masks in prompts, unifying visual chat with detection \\
 & Kosmos\textendash 2~\cite{peng2024grounding}      & ICLR'24   & Integrating ground tokens into vocabulary, allowing the model to generate bounding boxes as its instruction-driven output.\\
 & GLaMM~\cite{rasheed2024glamm} & CVPR'24 & Introducing mask token to vocabulary, enabling the model to predict the pixel-level mask  \\
 & ViP\textendash LLaVA~\cite{cai2024vip}     & CVPR'24       & Encoding visual instruction prompts via visual feature fusion \\
 & Draw\textendash and\textendash Understand~\cite{Lin2024DrawUnderstand} & ICLR'25 & Proposing a visual instruction prompt encoder \\
\cmidrule{2-4}
 & \multicolumn{3}{l}{\textit{\textbf{(e) Enhance output architecture.}}} \\
 & LISA~\cite{lai2024lisa}             & CVPR'24     & Introducing\texttt{<SEG>} token to generate decoder-based segmentation outputs \\
 & GSVA~\cite{xia2024gsva}             & CVPR'24        & Introducing \texttt{<SEG>} token for multi-targets mask and \texttt{<REJ>} token to reject empty targets \\
 
 & VisionLLM v2~\cite{Wu2024VisionLLMv2}     & NeurIPS'24       & Introducing unified multi\textendash task decoders \\
 
& VITRON~\cite{fei2024vitron} & NeurIPS'24 & Proposing an unified paradigm for video and image segmentation.\\
 & M2SA~\cite{Jang2025MMR} & ICLR'25   & Proposing early visual feature fusion and multiple \texttt{<SEG>} tokens to generate more fine-grained output\\
 & POPEN~\cite{zhu2025popen}            & CVPR'25        & Proposing preference-based segmention  \\
\midrule
\multirow{4}{*}{\shortstack{Dynamic\\Perception}}
 & \multicolumn{3}{l}{\textit{\textbf{(d) Visual search.}}} \\
 & V*~\cite{wu2024v}               & CVPR'24           & Iterative look\textendash back search \\
 & DyFo~\cite{li2025dyfo}            & CVPR'25           & MCTS\textendash guided focus \\
 & FaST~\cite{sun2024visual}             & ICLR'25       & Fast or slow visual search \\
\bottomrule
\end{tabular}
\label{tab:Task-Relevant Visual}
\vspace{-10pt}
\end{table*}

\subsection{Enhance Vision-Language Alignment in MLLMs}
\label{subsec:how-to-enhance-align}
While prior work has significantly improved the general-purpose representation capabilities of visual encoders, a new core focus has emerged in the field of MLLMs: \textbf{enhancing task-driven vision-language alignment for better interactive understanding.}
The core concept of this alignment is to effectively connect task-agnostic, general visual semantics, encompassing broad information about objects, attributes, and relations, with the specific goals and constraints contained in a user's instruction. This instruction-guided workflow transforms general-purpose visual representations into task-relevant ones through two primary stages: 
\begin{itemize}
\item \textit{\textbf{Task-Relevant Visual Representations Extraction.}} Conditioned on the instruction's semantics, the model selectively extracts relevant representations from the high-dimensional, general-purpose visual representation space (Sec. \ref{sec4.2.1}).
\item \textit{\textbf{Cross-Modal Fusion and Response Generation.}} The extracted visual representations are then deeply fused with the instruction's semantic representation to generate a precise, structured output that meets the task's requirements (Sec. \ref{sec4.2.2}).
\end{itemize}

However, the methods in these stages represent a static, single-turn form of interaction. To address this limitation, a key research direction has been to build a \textit{\textbf{dynamic perception}} mechanism (Sec. \ref{sec4.2.3}). The goal is to give the model the ability for active perception. Specifically, the model can evaluate its current understanding to determine if more visual evidence is needed, and then actively re-examine the image to get that evidence. This creates an iterative visual search loop.

\begin{figure}[t]
    \centering
    \includegraphics[width=1.0\linewidth]{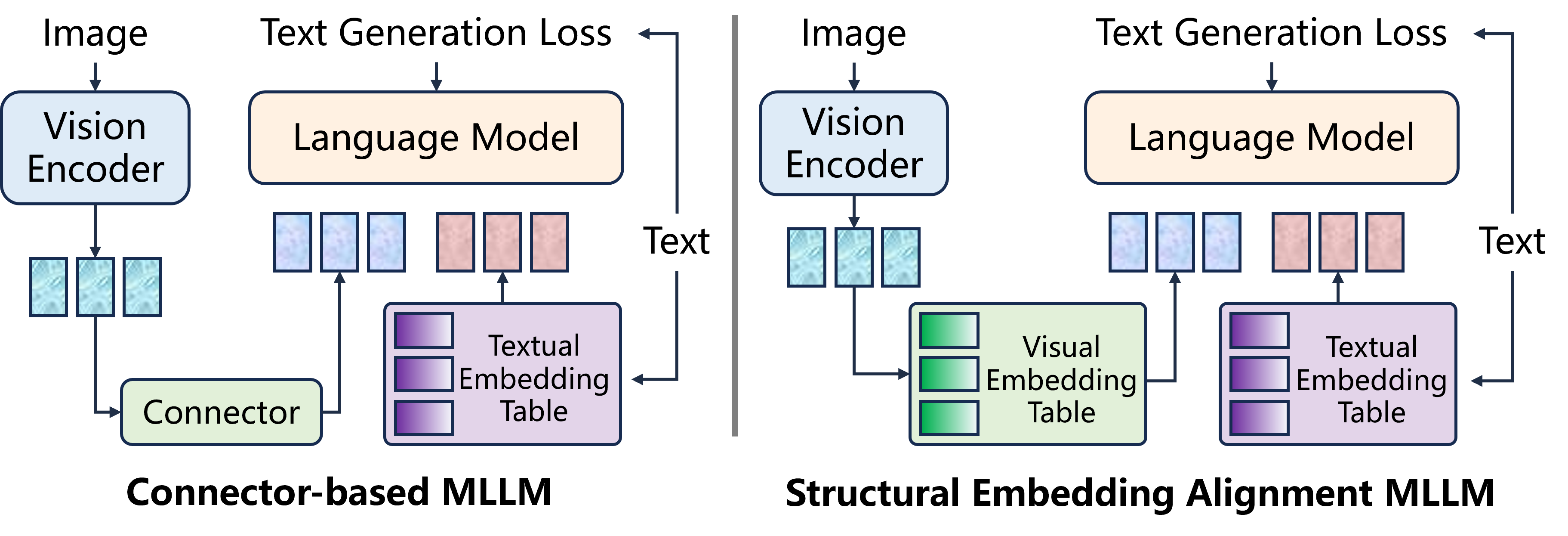}
    \vskip -0.15in
    \caption{Left: Connector-based MLLM: The typical architecture of traditional multimodal models (e.g., LLaVA), where the connector is usually an MLP that projects visual features into the same dimensional space as text embeddings. Right: Structured embedding alignment in Ovis: The output of the visual encoder is no longer directly projected through an MLP, but is instead mapped to a visual embedding table.}
    \label{figprojector}
\vskip -0.1in
\end{figure}

\subsubsection{Task-Relevant Visual Representation Extraction}
\label{sec4.2.1}
Following the aforementioned framework, we first explore the core strategies for the first stage: enhancing the model's ability to extract task-relevant visual representations. Researchers have primarily proposed three main approaches: \textbf{improving the projection layer}, \textbf{task-specific fine-tuning}, and \textbf{prompt tuning}. These methods focus on strengthening the expression of task-specific visual features by optimizing the alignment and adaptation mechanism that maps visual features into the language space, all without altering the underlying visual backbone. As illustrated in Table~\ref{tab:Task-Relevant Visual}, we categorize the representative works in task-relevant visual representation extraction.  

\noindent$\bullet$ \textit{\textbf{Improve the Projection Layer.}}
As the bridge connecting the visual encoder and the large language model, the design quality of the projection layer directly determines the depth and precision of the model's visual understanding. Traditional static projection methods, such as a simple MLP~\cite{taud2017multilayer}, indiscriminately map all visual features, creating a significant information bottleneck~\cite{li2023blip,dai2023instructblip,alayrac2022flamingo}. Therefore, the core significance of improving the projection layer lies not only in transmitting richer visual information but, more importantly, in empowering it with the ability to dynamically extract and transform relevant visual features based on the specific task.

For instance, in the fields of chart understanding and medical question answering, ChartMoE~\cite{xu2024chartmoe} and Uni-Med~\cite{zhu2024uni} employ a MOE~\cite{chen2022towards} architecture. This structure consists of multiple identical two-layer MLPs and uses a Top-k routing mechanism to weight and combine their outputs based on the input content. In video understanding, LLaVA-Octopus~\cite{zhao2025llava} adaptively fuses the outputs of multiple specialized projectors according to the given instruction. To enhance the model's fine-grained perception of spatial relationships, LLaVA-ST~\cite{li2025llava} and Honeybee~\cite{cha2024honeybee} respectively reinforce the spatial properties of visual features before projection by explicitly modeling local relationships and introducing space-aware convolutions. This is crucial for localization and referring tasks.
ParGo~\cite{wang2025pargo} proposes an innovative global-local projector that bridges vision and language by integrating both global context and local details, overcoming the over-focus on salient regions in traditional methods, enabling more comprehensive and fine-grained visual representation, while effectively controlling computational cost through token length management, thus achieving efficient alignment between visual features and the LLM.
As shown in Fig.\ref{figprojector}, the Ovis series~\cite{lu2024ovis,lu2025ovis2} no longer projects the output of the visual encoder through an MLP, but instead maps it into a learnable visual embedding table for transformation. This structure is similar to the text embedding table.

\noindent$\bullet$ \textit{\textbf{Task-Specific Fine-Tuning.}}
Instruction fine-tuning on task-specific vision-language datasets, such as LLaVA-Instruct-150K~\cite{liu2023visual} and MathV360K~\cite{shi2024math}, can enhance a model's ability to extract visual information tailored to those tasks. For example, MATCHA~\cite{liu2022matcha} focuses on mathematical reasoning; Q-Instruct~\cite{wu2024q} is dedicated to strengthening foundational visual perception for question answering; ChartGemma~\cite{masry2024chartgemma} and ChartInstruct~\cite{masry2024chartinstruct} concentrate on the comprehension of chart-based problems; and LLaVA-Med~\cite{li2023llava} targets visual question answering in the medical domain.

\noindent$\bullet$ \textit{\textbf{Prompt Tuning.}} 
Although instruction fine-tuning is effective, it has limited scalability and typically requires adjusting a large number of parameters, resulting in high training costs. To address this issue, Parameter-Efficient Fine-Tuning (PEFT) techniques have been introduced to the multimodal domain. The core idea originates from Visual Prompt Tuning (VPT)~\cite{jia2022visual} in the computer vision field. This method adds a small number of learnable prompt parameters to the input of a frozen backbone network, enabling adaptation to downstream tasks at a very low cost. Inspired by this, works such as VPGTrans~\cite{zhang2023vpgtrans}and TVP~\cite{zhang2024exploring} have designed visual prompts. Without compromising the model's general-purpose visual representations, these prompts enhance the alignment between task-specific semantics and visual features, thereby achieving efficient and transferable task adaptation.

\subsubsection{Cross-Modal Fusion and Response Generation}
\label{sec4.2.2}
\begin{figure}[t]
    \centering
    \includegraphics[width=1.0\linewidth]{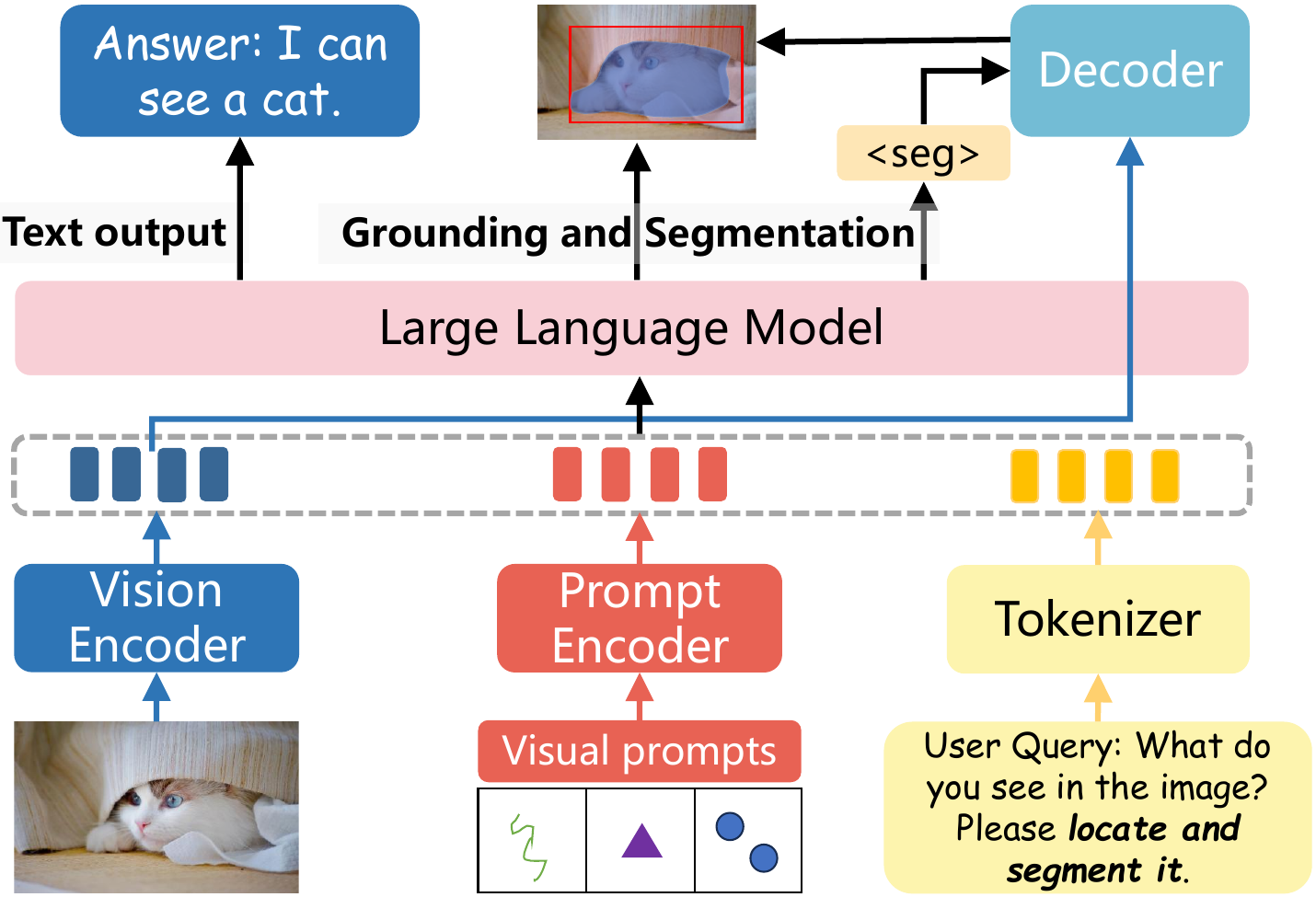}
    \vskip -0.1in
    \caption{An illustration of cross modal fusion and response generation. In this figure, the prompt encoder improves the instruction encoding paradigm. The segmentation decoder enables the model output the segmentation mask, which enhances the response generation.}
    \label{fig:fusion and generation}
\vskip -0.1in
\end{figure}

Once task-relevant visual representations are obtained, the subsequent challenge is to effectively align these representations with the semantics of the instruction to drive the generation of precise responses. In mainstream MLLM architectures, this cross-modal alignment process primarily relies on cross-attention mechanisms. Consequently, most current research is evolving along two parallel technical paths: The first path focuses on \textbf{improving the instruction encoding paradigm}, enabling the model to respond more precisely to specific semantic concepts within the instruction during the alignment process~\cite{zhang2024llava}. The second path, by \textbf{enhancing the output architecture}, aims to materialize the cross-modal understanding formed during fusion into responses that go beyond the scope of text~\cite{ren2024pixellm}. These two paths are complementary, sharing the common goal of achieving more fine-grained vision-language alignment. As depicted in Fig.~\ref{fig:fusion and generation}, we demonstrate the process of how to enhance the encode paradigm and response generation.

\begin{table*}[t!]
\footnotesize
\caption{An overview of CoT-focused training paradigms and the core issues they aim to address.}
\renewcommand\arraystretch{1}
\setlength{\tabcolsep}{6.5pt} 
\vspace{-5pt}
\centering
\begin{tabular}{l|l|l|p{9.2cm}}
\toprule
Method & Venue  & Training & Core Problem \\
\midrule
Let's Verify Step\textendash by\textendash Step~\cite{lightman2023let} & ICLR'24  & SFT & Outcome-based supervision cannot guarantee the logical correctness of the problem decomposition process itself. \\
Multimodal\textendash CoT~\cite{zhang2023multimodal} & ICLR'24 & SFT & Text-only reasoning decomposition decouples from visual facts. \\
ChartGemma~\cite{masry2024chartgemma} & ACL'24  & SFT & General MLLMs underperform on charts problems decomposition. \\

Dolphins~\cite{ma2024dolphins}  & ECCV'24  & SFT & Unlocking transferable reasoning capabilities for autonomous driving through fine-tuning. \\

Visual CoT~\cite{shao2024visual} & NeurIPS'24  & SFT & The individual steps of a problem decomposition are not explicitly grounded in verifiable local visual evidence.  \\

LLaVA\textendash CoT~\cite{xu2024llavacot} & arXiv'24  & SFT & Problem decomposition is an emergent ability from prompting, not an innate and robust skill of the model. \\

LlamaV-o1~\cite{thawakar2025llamav} & ACL'25  & SFT & Whether CoT enhances the adversarial robustness of MLLMs and how reasoning process behaves under adversarial attacks.\\

Learning Theorem Rationale~\cite{sheng2025learning} & AAAI'25  & SFT & When decomposing math problems, the reasoning processes of MLLMs fail to adhere relevant mathematical theorems. \\


UV\textendash DPO~\cite{zhao2025uvcot} & arXiv'25 & DPO & Learning to ground each step of a visual problem decomposition via SFT is data-hungry and costly. \\

Visual Grounded Reasoning~\cite{Wang2025VGR} & arXiv'25  & SFT &  The model's decomposition path is biased by language priors instead of being driven by fine-grained visual evidence. \\

VLM\textendash R1~\cite{shen2025vlmr1} & arXiv’25 & SFT+GRPO & SFT is insufficient for finding optimal decomposition policies\\
Visual\textendash RFT~\cite{liu2025visualrft} & arXiv’25  & SFT+GRPO &  It is difficult to define effective reward signals to guide the decomposition of complex visual problems via Reinforcement Learning. \\
\bottomrule
\end{tabular}
\label{tab:CoT-focused training}
\vspace{-10pt}
\end{table*}

\noindent$\bullet$ \textit{\textbf{Improve Instruction Encoding Paradigm.}}
To enhance the region-level referring capabilities of MLLMs, the instruction encoding paradigm has undergone a series of developments. Early models, such as Shikra~\cite{chen2023shikra} and ChatSpot~\cite{zhao2023chatspot}, directly encoded the continuous coordinates of a Region of Interest (ROI) as part of the text sequence. To facilitate model learning, Kosmos-2~\cite{peng2024grounding} first encodes the image after dividing it into $p \times p$ patches, and then discretizes and normalizes the continuous ROI coordinates, mapping them to positional tokens within the vocabulary. Subsequent research, including GLaMM~\cite{rasheed2024glamm} and Osprey~\cite{yuan2024osprey}, adopted binary masks as input, which provide greater spatial detail, to achieve more precise regional guidance. More recent research has shifted towards more flexible forms of visual instructions: ViP-LLaVA~\cite{cai2024vip} extended the input to include hand-drawn visual markers~\cite{shtedritski2023does}, while Draw-and-Understand~\cite{Lin2024DrawUnderstand} and OMG-Seg~\cite{Li2024OMGSeg} introduced an external visual prompt encoder. This allows the model to decode the user's intent from the visual channel without compromising its original global understanding capabilities.

\noindent$\bullet$ \textit{\textbf{Enhance the Output Architecture.}}
To reduce the ambiguity of text-only outputs in complex visual scenes, researchers have explored output architectures capable of generating pixel-level localization information. Models like LISA~\cite{lai2024lisa}, LLM-Seg~\cite{wang2024llm}, and OMG-LLaVA~\cite{zhang2024omg} introduced a special token \texttt{<SEG>}, into the vocabulary. The hidden state vector corresponding to this token is then used as a query to drive a separate segmentation decoder to generate a mask. This paradigm was further extended by works like GSVA~\cite{xia2024gsva}, MMR~\cite{Jang2025MMR}, Instruction-guided-masking~\cite{zheng2024instruction} to handle multi-object and multi-granularity segmentation tasks. To address the issue of semantic bias in segmentation results, POPEN~\cite{zhu2025popen} employed preference learning to fine-tune the model and suppress incorrect segmentations. Concurrently, to reduce the reliance on large-scale, pixel-level annotated data, works like Llafs~\cite{zhu2024llafs} and Prompt Highlighter~\cite{zhang2024prompt} have investigated instruction-based segmentation under few-shot or zero-shot conditions. Furthermore, several works~\cite{pi2024perceptiongpt,Wu2024VisionLLMv2,fei2024vitron,Shen_2024_CVPR} have focused on building a unified output interface. This enables a single model to generate different forms of localization results, such as bounding boxes or segmentation masks, according to the user's instruction, thereby accommodating diverse task requirements.

\subsubsection{Dynamic Perception}
\label{sec4.2.3}
Building upon the static, single-turn vision-language interaction paradigm discussed previously, this section focuses on the implementation of dynamic perception. The core of these methods is to endow the model with the ability to actively and iteratively search for visual information, thereby overcoming the limitations~\cite{lin2025mind} of static perception. 

For example, V*~\cite{wu2024v} employs an LLM-guided hierarchical visual search. At inference time, it collaborates with an MLLM to progressively zoom in and look back for evidence, achieving more fine-grained visual question answering in high-resolution and crowded scenes. Subsequent research has focused on reducing training costs and improving architectural coupling\cite{li2025dyfo},\cite{sun2024visual}, \cite{guo2025hierarchical}. DyFo~\cite{li2025dyfo} formalizes visual search as a Monte Carlo Tree Search (MCTS)~\cite{Browne2012MCTS}, interacting with external visual experts to dynamically focus on key regions in a training-free setting. Taking a step further, FaST~\cite{sun2024visual} , inspired by the concept of fast and slow thinking, trains a lightweight, built-in adapter to control the speed of reasoning based on the problem's difficulty, enabling human-like dynamic visual search.

\subsection{Enhance Problem Decomposition Ability of MLLMs}
\label{subsec:how-to-enhance-decompos}
Early MLLMs operated on a single-step reasoning paradigm, the core deficiency of which lies in treating any task as a monolithic "input-output" mapping process, without any explicit problem decomposition. Consequently, these models often falter when faced with complex reasoning tasks~\cite{Ma2023CREPE}. To overcome this limitation, research has focused on endowing models with the ability to perform step-by-step problem decomposition~\cite{zheng2023ddcot,Kojima2022ZeroShotReasoners}. The goal is not merely to improve the accuracy of the final answer, but also to ensure the correctness and verifiability of the reasoning process.

Presently, most approaches do not employ a built-in, structured framework for problem decomposition; instead, they rely on prompting techniques, such as Chain-of-Thought (CoT), to implicitly steer the model along a decompositional reasoning path. However, a sole reliance on prompting~\cite{zhang2023multimodal, wei2022chain} is insufficient for internalizing this decompositional ability within the model's parameters. The central research challenge is therefore to transform this decompositional reasoning from a transient, prompt-induced behavior into an innate, self-directed capability of the model.
To this end, the academic community~\cite{Wang2025MCoTSurvey} is primarily exploring the following three research directions: 
\begin{itemize}
\item \textit{\textbf{Enhancing problem decomposition via training.}} Dedicated training paradigms can enhance the problem decomposition ability of Multimodal Large Language Models (MLLMs), aiming to make their reasoning more visually grounded and logically sound. (Sec. \ref{subsubsec:4.3.1}).
\item \textit{\textbf{Automated Synthesis of Training Data.}} To overcome the high cost and scalability issues of manual annotation, researchers have developed automated methods for constructing large-scale, high-quality Chain-of-Thought (CoT) datasets with interleaved visual and textual evidence. (Sec. \ref{subsubsec:4.3.2}).
\item \textit{\textbf{Inference-Time Search for Flexible Decomposition.}} To overcome the limitations of traditional Chain-of-Thought (CoT), which follows a single, linear reasoning path and risks finding suboptimal solutions, researchers are adapting inference-time search algorithms to explore multiple reasoning paths and find the best possible answer. (Sec. \ref{subsubsec:4.3.3}).
\end{itemize}

\subsubsection{Enhancing Problem Decomposition via Training}
\label{subsubsec:4.3.1}
The paradigm of enhancing problem decomposition through process supervision originated in the language-only domain~\cite{lightman2023let}. When this paradigm is migrated to multimodal scenarios, researchers must address a series of new challenges. The primary issue is that the model's decomposition steps must be guided and constrained by visual information to ensure factual consistency. Secondly, for specialized domains such as mathematics, the model must acquire specific decomposition structures that align with the field's intrinsic logic. In Table~\ref{tab:CoT-focused training}, we present some representative methods to give a first-look understanding of prevailing training paradigms. These methods can be categorized into three main training paradigms:\textbf{imitation learning, curriculum learning and preference learning}.

\noindent$\bullet$ \textit{\textbf{Imitation Learning.}}
Early CoT methods established a foundational framework for problem decomposition, but their generated text-only reasoning paths were prone to decoupling from visual facts~\cite{wei2022chain}. As the entire reasoning process unfolded solely at the textual level, the decomposition plan could be flawed from the outset. Consequently, the core of subsequent research has been to introduce strict constraints to the decomposition process, ensuring that each step of the reasoning path and its required evidence remain consistent with the visual facts. Multimodal-CoT pioneered a two-stage generation process to achieve implicit supervision of visual evidence~\cite{zhang2023multimodal}. The model first generates a textual reasoning chain, which then serves as the condition for producing the final answer. This sequential dependency creates an implicit constraint, compelling the initial reasoning to be visually grounded to ensure the final answer's accuracy. Building on this, to enhance the verifiability of decomposition steps, works like Visual CoT and Visual Grounded Reasoning supervised the model to cite visual evidence such as bounding boxes, transforming ambiguous linguistic references into precise, verifiable coordinate localizations~\cite{shao2024visual, Wang2025VGR}. Furthermore, to improve the domain-specific decomposition logic of MLLMs, works such as ChartGemma~\cite{masry2024chartgemma}, Dolphins~\cite{ma2024dolphins},  ChartInstruct~\cite{masry2024chartinstruct}, Sce2DriveX~\cite{zhao2025sce2drivex} and Learning Theorem Rationale~\cite{sheng2025learning} have fine-tuned models on specialized CoT datasets. This ensures the model's reasoning process adheres to the rigorous paradigms of disciplines like mathematics, thereby validating the effectiveness of its decomposition.

\noindent$\bullet$ \textit{\textbf{Curriculum Learning.}}
More recently, LLaVA-CoT~\cite{xu2024llavacot} and LlamaV-o1~\cite{thawakar2025llamav} have proposed a ``de-prompted'' Supervised Fine-Tuning (SFT) paradigm that introduces the concept of curriculum learning to cultivate the model's decomposition and reasoning abilities through a phased, easy-to-hard process. The model's training unfolds in three stages: it first learns to decompose complex problems into sub-problems, then grounds each sub-problem in visual evidence, and finally integrates this evidence to form a coherent, summary answer. As a result, these models are trained to autonomously decompose problems at inference time and generate intermediate steps and the final answer accordingly. This curriculum-based training paradigm helps the model form a more robust and innate reasoning capability. However, the effectiveness of SFT is limited by its dependence on imitating a single "correct" decomposition path. This method does not equip models to evaluate alternative strategies or self-correct their reasoning, which limits their generalization in tasks that require flexible, exploratory reasoning. However, the effectiveness of SFT is limited by its dependence on imitating a single "correct" decomposition path. This method does not equip models to evaluate alternative strategies or self-correct their reasoning, which limits their generalization in tasks that require flexible, exploratory reasoning.

\noindent$\bullet$ \textit{\textbf{Preference Learning.}}
The core idea of preference learning is to learn a predictive model from feedback expressed as relative preferences. This model aims to capture and internalize the underlying judgment criteria hidden within the feedback, thereby enabling it to make preference judgments on unseen instances that align with those criteria. Unlike traditional imitation learning, which provides a positive label for each sample, datasets for preference learning are composed of preference relationships in forms such as comparisons, rankings, or ratings. From a policy optimization perspective, mainstream methods can be divided into on-policy and off-policy learning. 

\textbf{For on-policy learning,} the core principle is that the data used for policy updates must be generated by the current version of the policy being optimized. This paradigm ensures an unbiased update direction and is generally more stable. Mainstream methods adopt Generalized Reward Policy Optimization~\cite{shao2024deepseekmath} (GRPO) and its variants. For example, methods like VLM-R1~\cite{shen2025vlmr1}, Visual-RFT~\cite{liu2025visualrft}, Reason-RFT~\cite{tan2025reasonrft}, Seg-Zero~\cite{liu2025segzero}, R1-OneVision~\cite{yang2025r1} and RAGEN~\cite{wang2025ragen} transform visual evidence from the reasoning process into verifiable training signals, thereby enhancing the model's ability to select better problem decomposition paths. 

\textbf{For off-policy learning,} the core principle is that policy updates can utilize historical experience data generated by any previous policy. It significantly improves data efficiency by correcting for distribution shifts using techniques like importance sampling. Mainstream methods adopt DPO~\cite{rafailov2023dpo} (Direct Preference Optimization) and its variants. For instance, UV-CoT~\cite{zhao2025uvcot} generates multiple chains of thought for the same visual problem, constructs preference pairs based on visual alignment, step-by-step consistency, and the correctness of the final answer, and then updates the policy. V-DPO~\cite{xie2024v} uses a large number of synthetically generated image preference pairs to enhance the model's inclination to follow visual evidence when decomposing problems during reasoning. VTS-DPO~\cite{bai2025multi} trains a general-purpose verifier/scorer offline using preference pairs from multi-step trajectories to stably evaluate and guide multi-step visual reasoning at inference time. 

\begin{figure*}[t]
    \centering
    \includegraphics[width=1.0\linewidth]{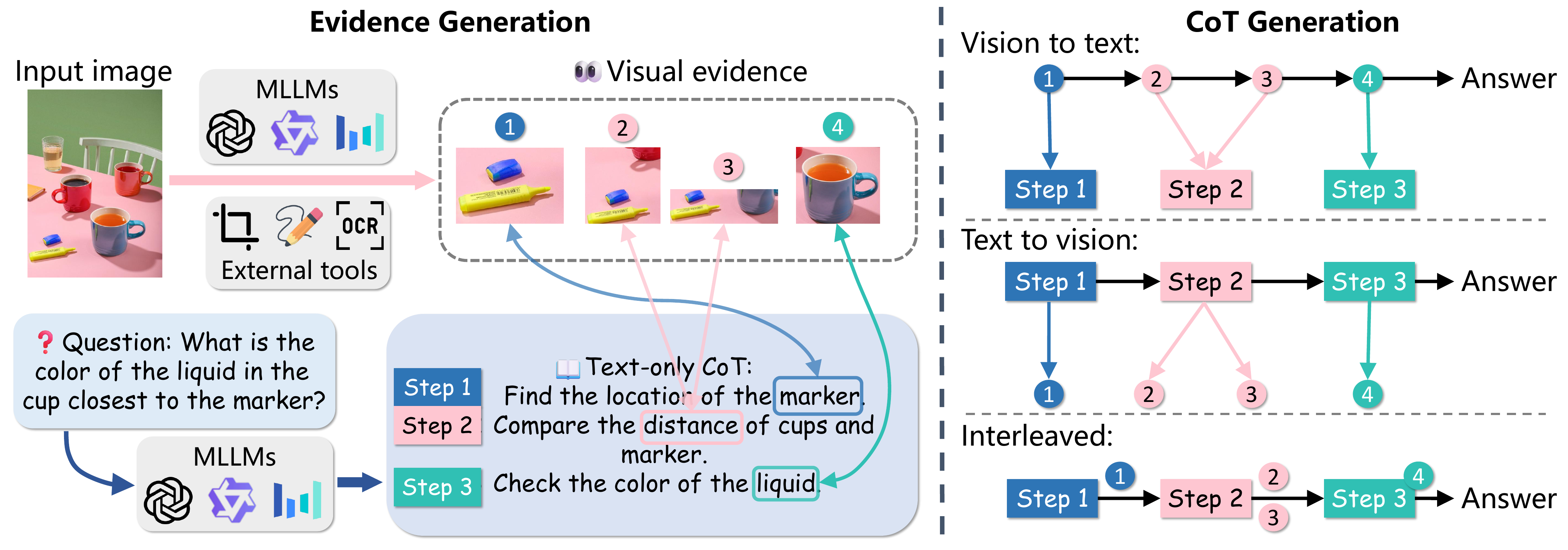}
    \vskip -0.15in
    \caption{An example of how to construct an interleaved CoT. We demonstrate three typical ways in the image. The main difference is that, interleaved method generates text rationales and visual evidence in a coherent way, while another two  generate text and visual evidence respectively.}
    \label{fig:datacuration}
\end{figure*}

\subsubsection{Automated Synthesis of Training Data}
\label{subsubsec:4.3.2}
The effectiveness of supervised learning methods in enhancing a model's problem decomposition ability is highly dependent on large-scale, high-quality CoT datasets. Early benchmarks, such as ScienceQA~\cite{lu2022scienceqa} and A-OKVQA~\cite{schwenk2022okvqa}, provided detailed, manually annotated reasoning steps and answers for each problem. However, manual annotation is costly, has limited scalability, and struggles to meet the data volume requirements of current training paradigms. Furthermore, this approach often lacks visual evidence grounding for the intermediate steps, making it ineffective at enhancing a model's ability to decompose specifically visual problems. To mitigate these issues, researchers have adopted automated data synthesis methods. It is important to note that ``synthesis" here is defined broadly, encompassing both the generation of entirely new CoT~\cite{wei2022chain} data from scratch for specific tasks, as well as the reprocessing and enhancement of existing datasets which mainly rely on augmenting text-only reasoning processes with visual evidence. Based on the methods for generating and processing CoT data, these approaches can be categorized into two primary strategies: \textbf{generation via external teacher models} and \textbf{bootstrapped data generation}.

\noindent$\bullet$ \textit{\textbf{Generation via External Teacher Models.}}
This paradigm primarily employs a teacher model to automate the generation of CoT datasets. Its core task is to decompose a vision-language reasoning problem step-by-step, providing both the correct reasoning process and the final answer. However, the key challenge lies in ensuring that the decomposition path generated by the model is not only logically sound at the textual level but also tightly aligned with the visual evidence. To address this challenge, researchers have explored various synthesis strategies. As illustrated in Fig.~\ref{fig:datacuration}, these strategies can be categorized into three mainstream approaches based on the sequence in which the textual decomposition steps and the visual evidence are generated and verified: \textbf{Vision-to-Text}, \textbf{Text-to-Vision} and \textbf{Interleaved Vision-Text Generation}.

\textbf{(a) Text-to-Vision.}
The Text-to-Vision (T2V) paradigm first constructs a text-only reasoning sequence and subsequently matches each textual reasoning step with corresponding visual evidence to build the interleaved reasoning dataset. For example, Cogcom~\cite{qi2024cogcom} first uses a teacher model to generate a reasoning process that contains placeholders for visual operations. It then executes these operations to populate the visual results and searches the resulting branching tree with a DFS~\cite{tarjan1972depth} algorithm to retrieve the optimal reasoning path. The data construction process for Mint-CoT~\cite{chen2025mintcot} is based on the existing Mulberry-260K~\cite{yao2024mulberry} dataset. It begins by filtering for mathematical problems and their text-only solutions. Next, it utilizes Optical Character Recognition (OCR) to extract key visual tokens from the corresponding images. Finally, a teacher model is employed to precisely align each textual reasoning step with the visual tokens upon which it depends. However, in this construction paradigm, the generation processes for textual reasoning and visual evidence are entirely decoupled, which can easily introduce hard-to-correct visual hallucinations at the source.

\textbf{(b) Vision-to-Text.}
This paradigm employs a \textbf{vision-first} approach, where a textual rationale is generated only after visual evidence has been established. Representative works like MM-GCoT~\cite{wu2025mmgcot} and SIFThinker~\cite{chen2025sifthinker} simulate the human viewing pattern by proceeding from a global scan to local details to identify and structure this evidence. Similarly, Pixel-Reasoner~\cite{su2025pixelreasoner} first uses a teacher model to locate key visual cues and obtain local views before generating its analysis. In these methods, the process is an evidence-driven "reverse induction": the textual reasoning is retroactively generated to describe and connect the pre-selected visual cues. While this approach effectively eliminates factual hallucinations by design, its fundamental drawback is that the resulting decomposition is not a top-down problem-solving strategy. Instead, its structure is dictated by the sequence of visual observations, which can lead to a reasoning path that is unnatural, inefficient, or logically suboptimal, despite being factually correct at every step.

\textbf{(c) Interleaved Vision-Text Generation.}
The joint synthesis of textual reasoning and visual evidence represents a less prevalent research direction. For instance, approaches like LATTE~\cite{ma2024latte} and TACO~\cite{ma2024taco} adopt a collaborative paradigm between a teacher model and multiple external visual tools, in which the teacher model is responsible for decomposing complex problems and planning the reasoning steps. When needed, it calls upon specific visual tools to acquire precise visual evidence. The results returned by these tools are then fed back into the reasoning process, creating a ``think-act-observe" loop that ultimately generates an interleaved vision-language Chain-of-Thought.

\begin{figure}[t]
    \centering
    \includegraphics[width=1.0\linewidth]{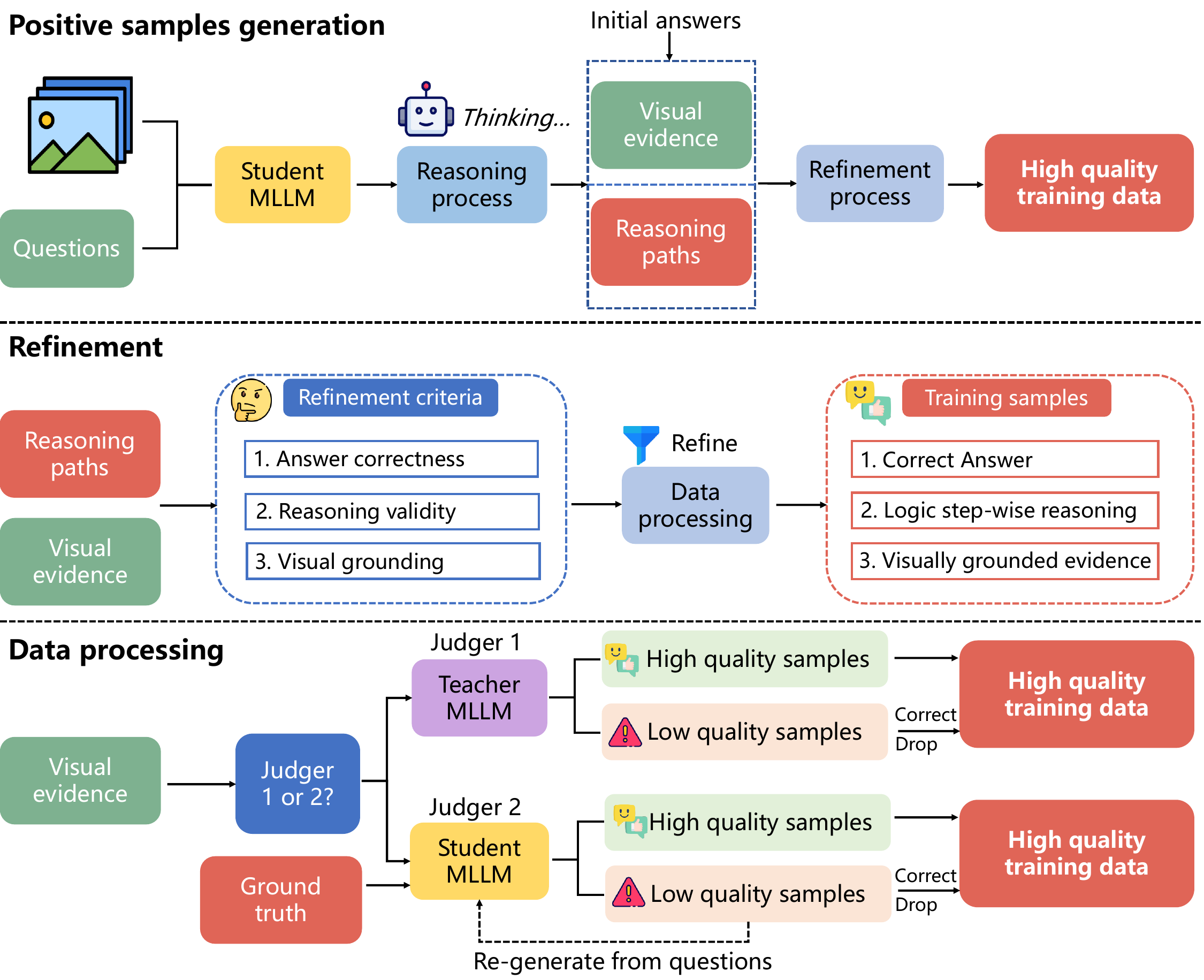}
    \vskip -0.15in
    \caption{An illustration of positive samples refinement.}
    \label{fig:bootstrap}
\end{figure}

\noindent$\bullet$ \textit{\textbf{Bootstrapped Data Generation.}}
To reduce the reliance on external teacher models, some works employ bootstrapping methods to generate CoT data. The core mechanism is to leverage the model's own capabilities to generate diverse reasoning paths for a given problem. These paths are then classified as positive or negative samples. This classification is based on the quality of their final outcomes or the logical validity of their reasoning process, with the evaluation performed either by the model itself or by external models. To ensure a rich set of negative examples, some methods also proactively generate them by intentionally injecting errors into the input or the reasoning chain. Through training on such curated data, the model learns to discriminate among candidate reasoning trajectories and prefer those yielding correct, evidence-aligned solutions. Based on whether the generated training data includes negative samples, we classify the methods into two categories: \textbf{Positive Sample Refinement} and \textbf{Preference Data Generation}. As illustrated in Fig.~\ref{fig:bootstrap}, we demonstrate the generation of positive samples.

\textbf{(a) Positive Sample Refinement.}
In the text-only domain, early research such as STaR~\cite{zelikman2022star} proposed a method for dynamically augmenting datasets. In this process, the model first generates decomposition paths for a problem based on few-shot examples. If a path proves effective and yields the correct answer, it is considered a high-quality training sample. If not, the model is guided to reverse-engineer a rational decomposition strategy from the correct answer. These corrected and validated high-quality samples are then fed back into the training set, creating a loop of continuous dataset optimization and simultaneous improvement of the model's capabilities. 

In the multimodal domain, this dataset construction strategy has been adapted to generate more complex reasoning samples that contain visual evidence. For instance, MC-CoT~\cite{tan2023boosting} constructs training data via bootstrapped filtering to improve the consistency~\cite{wang2022self} of smaller models in problem decomposition. It first prompts the model to self-iterate and generate multiple candidate decomposition paths. It then uses self-evaluation and a consistency voting mechanism to filter for the solution with the strongest consensus, which is then used as a positive sample for contrastive training against the correct label. GCOT~\cite{xia2025bootstrapping} demonstrates how to construct an interleaved vision-language Chain-of-Thought dataset with limited data. This method begins by decomposing a complex problem into a series of initial "sub-problem, bounding box" pairs to form a seed dataset. The model then trains on this dataset while continuously filtering and optimizing, retaining only the bounding boxes that best match the visual facts, ultimately producing a complete interleaved CoT dataset.

\begin{figure}[t]
    \centering
    \includegraphics[width=1.0\linewidth]{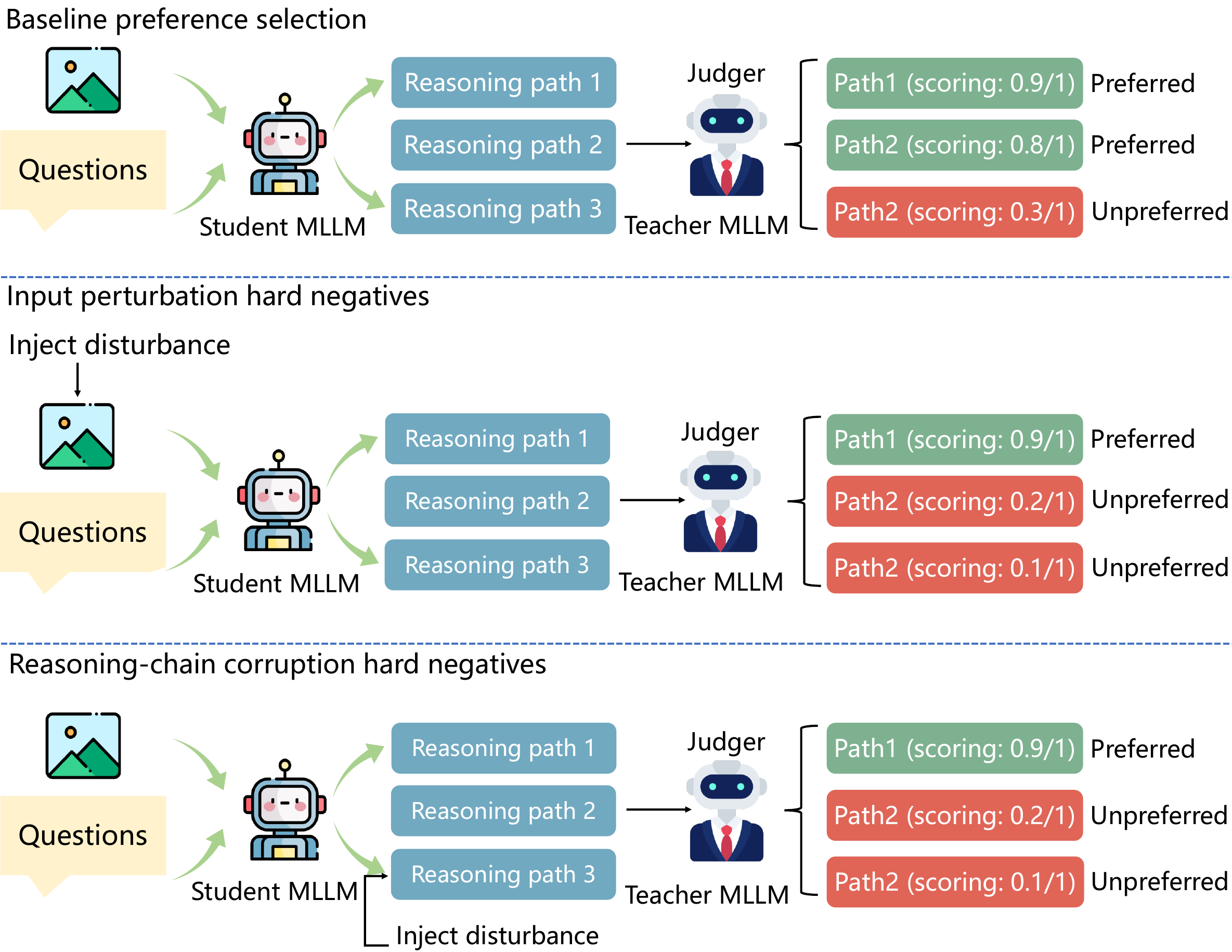}
    \vskip -0.15in
    \caption{An example of how to construct preference data. We obtain negative training samples by injecting perturbations into either the inputs or the candidate CoT outputs; alternatively, negatives can be collected without explicit perturbations.}
    \label{fig:preference data generation}
\vskip -0.1in
\end{figure}

\begin{figure*}[t]
    \centering
    \includegraphics[width=1.0\linewidth]{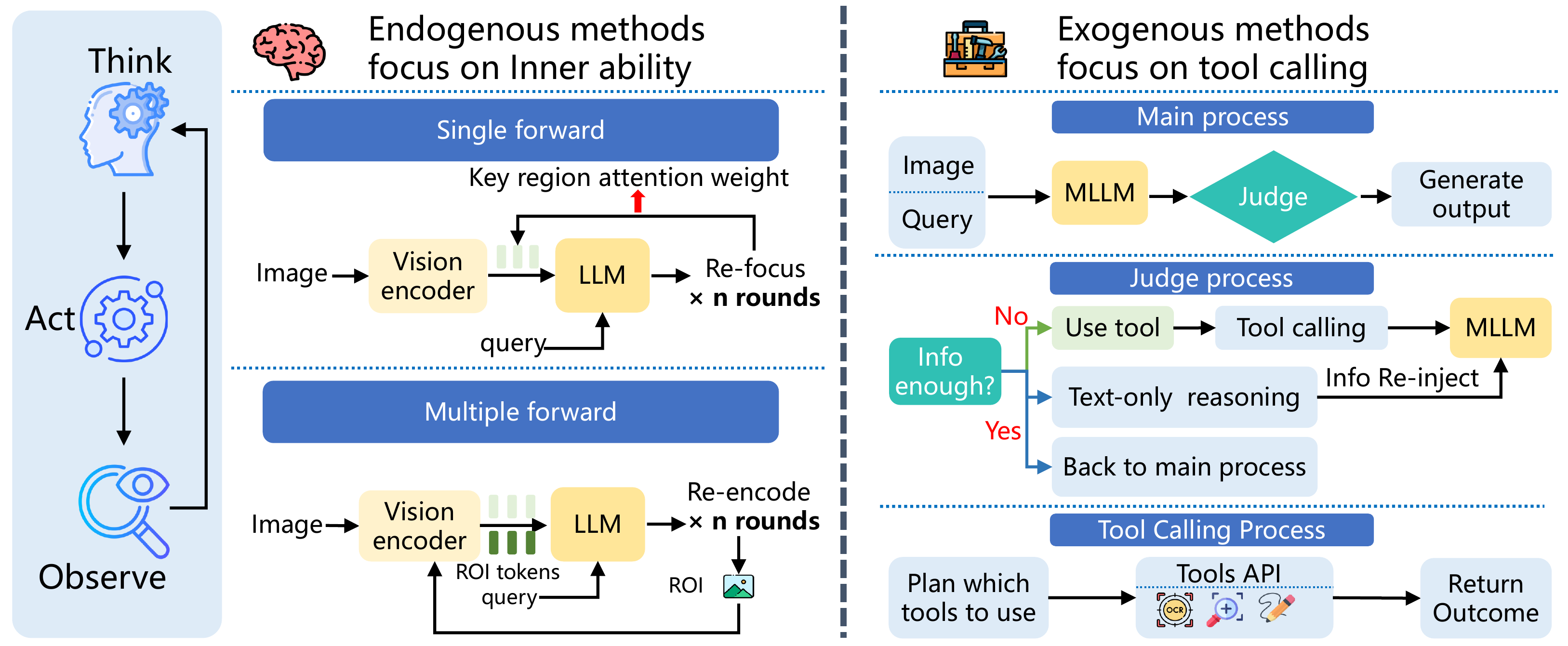}
    \vskip -0.15in
    \caption{An illustration of thinking with image.}
    \label{fig:think with image}
\vskip -0.1in
\end{figure*}

\textbf{(b) Preference Data Generation.}
Beyond generating only positive samples, other works focus on the automatic construction of preference pairs for offline learning methods like Direct Preference Optimization (DPO)~\cite{rafailov2023dpo}. As is depicted in Fig.~\ref{fig:preference data generation}, we give an example of how to construct preference data. This is primarily achieved through two strategies: judge-based scoring and bootstrapping. The former uses an external or base model as a ``judge'' to score multiple candidate responses. This is implemented by either evaluating the final outcomes of different reasoning paths~\cite{yu2024rlaifv,zhang2024improve,yang2023rlcd} or by using the model's capacity for self-correction to create positive and negative pairs~\cite{he2024self}. The latter strategy, bootstrapping, proactively generates negative samples by injecting perturbations. For example, BPO~\cite{pi2024strengthening} perturbs the image input or injects logical errors into the reasoning chain to automatically create valuable negative samples.

\subsubsection{Inference-Time Search for Flexible Decomposition}
\label{subsubsec:4.3.3}
The traditional CoT paradigm guides a model along a single, linear reasoning path. This process is essentially a greedy search strategy, which risks converging on a locally optimal path while missing the globally optimal solution. To overcome this limitation, methodologies originating from the text-only domain, designed to enhance the flexibility of the reasoning process, are being progressively adapted for multimodal tasks.

These methods replace the single, deterministic generation process by exploring a broader solution space. Foundational work such as Tree of Thoughts (ToT)~\cite{yao2023tree}  conceptualizes the reasoning process as a tree, where each node represents a partial solution or an intermediate thought. It systematically explores this tree using algorithms like Breadth-First Search (BFS)~\cite{bundy1984breadth} or Depth-First Search (DFS), enabling backtracking and the exploration of multiple reasoning paths. Building on this foundation, a series of advanced decoding strategies~\cite{wang2022self,gui2024bonbon,xie2023self,holtzman2019curious} have been proposed, which employ diverse sampling and voting mechanisms to identify the most robust answer. More recently, techniques like AlphaMath~\cite{chen2024alphamath}, Rest-mcts*~\cite{zhang2024rest}, and SVPO~\cite{chen2024step} model the reasoning process as a Monte Carlo Tree Search (MCTS) problem, allowing the model to dynamically allocate more computational resources to the most promising branches of the reasoning tree.

Although these advanced search algorithms have achieved success in unimodal tasks, their application in the multimodal domain is still in its nascent stages. Recent research has begun to bridge this gap, VisuoThink~\cite{wang2025visuothink} extends the foundational ToT algorithm into a forward-looking tree search algorithm. Meanwhile, Socratic-MCTS~\cite{acuna2025socratic}, vrest~\cite{zhang2025vrest} and A star~\cite{wu2025boosting} have adapted MCTS to the vision-language domain. These methods model the visual question answering task as the construction of a reasoning tree composed of nodes representing (sub-question, sub-answer) tuples. A key innovation in this adaptation is the use of the model's own self-consistency score with respect to visual evidence as a reward signal to guide the tree search. This promotes the generation of more visually-grounded reasoning paths.

\subsection{Dynamic Forensics During Reasoning}
\label{subsec:how-to-mitigate-hallucinations}

MLLMs typically employ a static, single-pass visual encoding mechanism when processing complex vision-language reasoning tasks, converting the image into a fixed representation before inference begins. However, a complex reasoning process demands that the model dynamically refocus its attention on the most critical visual evidence based on its evolving inferential needs. The existing static mechanism restricts this dynamic interaction, making it prone to two critical flaws: first, the initial visual information tends to decay as text generation progresses. Secondly, the model may over-rely on its linguistic priors, generating hallucinations~\cite{huang2024opera,li2025mitigating} that contradict the visual facts.

The core solution is to establish a ``think with image"~\cite{su2025thinking} reasoning loop, thereby enabling the model to continuously revisit visual evidence during inference. As illustrated in Fig.~\ref{fig:think with image}, we illustrate the Think–Act–Observe reasoning loop. This section focuses on the central strategy for implementing such a loop: constructing interleaved vision-language Chains-of-Thought. Based on the source of the visual evidence used in this construction, the approaches can be classified into two main categories: 

\begin{itemize}
\item \textit{\textbf{Vision-Language Alignment via Endogenous Visual Evidence Injection.}} To enable a model to dynamically refocus on visual evidence during reasoning without using external tools, researchers have developed endogenous methods that leverage the model's internal mechanisms. (Sec. \ref{subsubsec:4.4.1}).
\item \textit{\textbf{Vision-Language Alignment via Exogenous Visual
Evidence Injection.}} Exogenous methods treat the Multimodal Large Language Model (MLLM) as an intelligent agent that actively gathers visual evidence by calling external tools, creating an "observe-act-decide" reasoning loop.(Sec. \ref{subsubsec:4.4.2}).
\end{itemize}

\subsubsection{Vision-Language Alignment via Endogenous Visual Evidence Injection}
\label{subsubsec:4.4.1}

\begin{table*}[t!]
\setlength\tabcolsep{4pt}
\footnotesize
\caption{An overview of representative methods about endogenous visual evidence injection.}
\renewcommand\arraystretch{1}
\setlength{\tabcolsep}{13pt} 
\vspace{-5pt}
\centering
\begin{tabular}{l|l|l|p{9.2cm}}
\toprule
Method & Venue & Granularity & Main Contribution \\
\midrule
 & \multicolumn{3}{l}{\textit{\textbf{(a) Single forward pass.}}} \\
CVC~\cite{hu2025boosting}          & IJCAI'25  & token            & Proposing a mask-then-predict strategy that re-weights attention toward salient visual cues when needed. \\
ICoT~\cite{gao2025interleaved}         & CVPR'25   & bbox             & Selecting the most relevant regions via attention maps and re-encodes the corresponding crops. \\
MINT-CoT~\cite{chen2025mintcot}     & arXiv'25  & token            & Before each reasoning step, identify the most relevant visual tokens and re-inject them to guide generation. \\
Look-back~\cite{yang2025look}    & arXiv'25  & token            & Invoking an internal look-back mechanism to revisit previously attended visual cues when inconsistency is detected. \\
\cmidrule{1-4}
& \multicolumn{3}{l}{\textit{\textbf{(b) Multi forward passes.}}} \\
CogCoM~\cite{qi2024cogcom}       & ICLR'25   & bbox             & Introducing chain-of-manipulations reasoning that iteratively localizes, acts on, and verifies regions. \\
DeepEyes~\cite{zheng2025deepeyes}     & arXiv'25  & bbox             & Using end-to-end RL to decide when to re-encode task-critical regions. \\
Pixel Reasoner~\cite{su2025pixelreasoner} & arXiv'25 & bbox            & Training with SFT+RL to plan and execute visual operations and iteratively refine evidence. \\
SIFThinker~\cite{chen2025sifthinker}   & arXiv'25  & bbox + depth     & Simulating human visual search and augmenting with depth cues to strengthen spatial reasoning. \\
CMMCoT~\cite{zhang2025cmmcot}       & arXiv'25  & bbox             & Maintaining a memory bank to aggregate cross-image evidence for multi-image reasoning, revisiting key regions as needed. \\
\bottomrule
\end{tabular}
\label{tab:endogenous visual}
\vspace{-3pt}
\end{table*}

Endogenous methods aim to refocus on relevant visual information through the model's internal attention mechanisms, without relying on external tools. Based on the coupling between the reasoning process and the visual encoder, these methods can be classified into two categories: those involving a \textbf{single forward pass} and those involving \textbf{multiple forward passes}. In Table~\ref{tab:endogenous visual}, we provide an overview of the methods that takes this paradigm.

\textbf{A single forward pass} approach involves the encoder performing one global encoding of the visual input prior to inference. During the subsequent text generation and reasoning phase, the model dynamically assigns higher attention weights to critical visual information based on the current reasoning context, thereby integrating visual evidence into the process. In contrast, a \textbf{multiple forward passes} approach allows the model to actively request a re-encoding of specific visual regions based on intermediate reasoning results. This enables it to obtain higher-resolution or more targeted features on an on-demand basis.

\noindent$\bullet$ \textit{\textbf{Single Forward Pass.}}
Initially, MM-COT~\cite{zhang2023multimodal} proposed a method to simulate refocusing. It first takes the image and the initial question as input to generate a Chain-of-Thought text containing key visual evidence, which is then concatenated back with the original input to derive the final answer. However, this approach suffers from a rigid interaction pattern and is computationally expensive. To address this issue, subsequent works have adopted strategies based on implicit and explicit attention guidance. \textbf{Implicit attention guidance} aims to redistribute attention through internal mechanisms without explicit visual cues; for instance, Look-back~\cite{yang2025look} achieves an implicit ``re-look" by dynamically adjusting attention weights on key objects during text generation, while CVC~\cite{hu2025boosting} masks specific image regions to compel the model to backtrack and use other visual cues. \textbf{Explicit attention guidance}, in contrast, relies on clearly defined visual evidence. Point-RFT~\cite{ni2025point}, for example, uses bounding box coordinates to direct the model's attention to a specific area, whereas other works like Don't Look Only Once~\cite{chung2025don}, MINT-CoT~\cite{chen2025mintcot} and ICOT~\cite{gao2025interleaved} dynamically select the most relevant visual tokens from global features to aid the current reasoning step.

\noindent$\bullet$ \textit{\textbf{Multiple Forward Passes.}}
To obtain more reliable visual evidence, a line of research employs the strategy of multiple forward passes. Initial methods like Textcot~\cite{luan2024textcot} follow a sequential ``describe-then-crop" paradigm, which generate a detailed image description to guide the cropping and reasoning over these image areas. However, this approach is prone to the loss of fine-grained visual details and suffers from an inflexible reasoning process. To overcome this, Studies such as CMMCoT~\cite{zhang2025cmmcot} realize ``think with image" paradigm by dynamically interleaving visual grounding with the ongoing reasoning process. This is achieved by \textbf{adaptively} re-encoding key regions of interest (RoIs) at crucial steps and injecting these updated visual features into the subsequent reasoning chain. However, when to seek for more visual information is still an open question for community. Currently, some works like Sketchpad adopt in-context learning to follow prompts~\cite{hu2024visual-sk}, works such as Chain-of-spot~\cite{liu2024chain}, PromViL~\cite{le2025progressive}, Vocot~\cite{li2024vocot} adopt supervised fine-tuning to imitate expert paths, and works including Deepeyes~\cite{zheng2025deepeyes}, PixelReasoner~\cite{su2025pixelreasoner}, Active-O3~\cite{zhu2025active} and VLM-R$^3$~\cite{jiang2025vlm} adopt reinforcement learning to autonomously discover the optimal timing.

Further advancements build upon this paradigm, enhancing the quality and granularity of visual information acquired during reasoning. For instance, CogCoM~\cite{qi2024cogcom} broadens the visual action space into a composable ``chain of operations" like localization and labeling. Meanwhile, to improve spatial understanding, SIFThinker~\cite{chen2025sifthinker}, PixelThink~\cite{wang2025pixelthink} and LLaVA-Aurora~\cite{bigverdi2025perception} introduce depth information to enable joint object-level and spatial perception. This line of research has culminated in unified frameworks, such as the ``RoI re-encoding and visual token re-sampling" proposed by Argus~\cite{man2025argus}, which significantly enhances visual grounding while maintaining reasoning efficiency.

\begin{table*}[t!]
\setlength\tabcolsep{4pt}
\footnotesize
\caption{An overview of representative methods about exogenous visual evidence injection.}
\vspace{-5pt}
\centering
\begin{tabular}{l|l|l|p{9.2cm}}
\toprule
Method & Venue & Tools & Main Contribution \\
\midrule

& \multicolumn{3}{l}{\textit{\textbf{(a) In Context Learning.}}} \\
MM-ReAct~\cite{yang2023mmreact}         & arXiv'23   & Visual experts                & Proposing calling visual tools whenever a reasoning step requires additional evidence. \\
ViperGPT~\cite{suris2023vipergpt}         & ICCV'23    & GLIP/X-VLM/MiDaS              & Introducing a program-as-agent paradigm that generates executable code to invoke visual tools. \\
CLOVA~\cite{gao2024clova}            & CVPR'24    & OWL-ViT/BLIP/CLIP             & Introducing learnable prompts to select and adapt tool usage during reasoning. \\
Visual Sketchpad~\cite{hu2024visual-sk} & NeurIPS'24 & SAM/G-DINO/Matplotlib         & Mimicing human problem solving by sketching while reasoning.\\

\cmidrule{1-4}
& \multicolumn{3}{l}{\textit{\textbf{(b) Fine Tuning.}}} \\
LLaVA-Plus~\cite{liu2024llava}       & ECCV'24    & OCR/SAM/BLIP2                 & Training on tool-grounded manipulation traces, improving the model’s ability to use tools. \\
TACO~\cite{ma2024taco}             & ICLR 2025  & DepthAnything/OCR/SAM         & Invoking external tools when a step requires additional evidence or verification.  \\
OpenThinking~\cite{su2025openthinkimg}     & arXiv'25   & Crop/Point/Zoomin             & Proposing a unified API for visual operations to standardize and streamline tool invocation. \\
 
\bottomrule
\end{tabular}
\label{tab:exogenous visual}
\vspace{-3pt}
\end{table*}

\subsubsection{Vision-Language Alignment via Exogenous Visual
Evidence Injection}
\label{subsubsec:4.4.2}

Exogenous methods frame the MLLM as an intelligent agent that dynamically acquires visual evidence by calling external tools or interacting with an environment, creating an ``observe-act-decide" reasoning loop. This paradigm has evolved from an early ``plan-then-execute" model to a more flexible, interleaved Chain-of-Thought approach. In Table~\ref{tab:exogenous visual}, we provide an overview of the methods that takes this paradigm.

Pioneering works such as Visual Programming~\cite{Gupta_2023_CVPR}, ViperGPT~\cite{suris2023vipergpt}, HuggingGPT~\cite{shen2023hugginggpt} and InternGPT~\cite{liu2023interngpt} established a tool-assisted reasoning paradigm where the model performs a one-shot decomposition of a complex problem into a sequence of tool calls, often guided by in-context learning (ICL)~\cite{dong2022survey}. An executor then runs this pre-defined plan to generate a final answer. While this approach offers a clear planning path, its rigidity and lack of dynamic feedback make it brittle in complex, multi-step interactions and highly dependent on prompt engineering. To overcome this rigidity, subsequent research has shifted towards an adaptive, interleaved Chain-of-Thought model. In this paradigm, the model makes a decision at each reasoning step, selecting and executing a relevant visual operation and then using the immediate observation to inform its next action. The model acquires this step-by-step reasoning capability through either \textbf{in-context learning} or \textbf{fine-tuning}.

\noindent$\bullet$ \textit{\textbf{In-Context Learning.}}
Early works like ReAct~\cite{yao2023react} and MM-ReAct~\cite{yang2023mmreact}, along with later approaches such as Visual Sketchpad~\cite{hu2024visual-sk}, primarily rely on the in-context learning (ICL) capabilities of powerful foundation models, using meticulously designed prompts to guide tool use. While this method is highly flexible, it places stringent demands on the model's instruction-following capabilities and depends heavily on high-quality prompt engineering. Consequently, its stability is often challenged in complex tasks. To improve the robustness of ICL, subsequent research like CLOVA~\cite{gao2024clova} and VisRep~\cite{khan2024self} has introduced execution feedback and self-correction mechanisms. 

\noindent$\bullet$ \textit{\textbf{Fine-Tuning.}}
To internalize tool-use capabilities as an intrinsic skill, a significant body of research has employed fine-tuning. For instance, works like LATTE~\cite{ma2025latte}, LLaVA-Plus~\cite{liu2024llava}, DWIM~\cite{ke2025dwim}, Refocus~\cite{fu2025refocus} and VPD~\cite{hu2024visual} use supervised fine-tuning on tool-use trajectory data to teach the model when and how to use tools. Building upon this, MLLM-Tool~\cite{wang2025mllm} expands the modality into audio. In pursuit of more optimal decision-making policies, VTool-R1~\cite{wu2025vtool} and Visual-ARFT~\cite{liu2025visual} introduce reinforcement learning, training the model to master the best timing and sequence for tool calls through trial and error. Furthermore, to enhance the generality and usability of this paradigm, researches such as OpenThinkimg~\cite{su2025openthinkimg} has encapsulated diverse visual tools into a unified API, allowing the model to dispatch them on-demand during inference.

In summary, this section has systematically reviewed the cutting-edge methodologies for enhancing the interactive reasoning capabilities of MLLMs, structured along our proposed ``From Perception to Cognition" framework. At the \textbf{perceptual level}, the research has progressed along two major trends: one involves adopting dynamic strategies that integrate multiple expert encoders to capture richer, multi-granularity features; the other focuses on unifying image generation and understanding tasks, leveraging the powerful representational capabilities of generative models to enhance the perceptual foundation.

At the \textbf{cognitive level}, the focus has shifted from initial reliance on prompt-engineered, single-path imitation learning to a more systematic, multi-paradigm fusion. This includes supervised fine-tuning on visually-grounded Chain-of-Thought data, employing preference learning (e.g., DPO/GRPO) to select among multiple candidate paths, and utilizing tree-based search at inference time to explore a more optimal solution space. This convergence of methods aims to endow models with greater autonomy in problem decomposition and dynamic verification.

This synergistic development at both the perceptual and cognitive levels has collectively driven significant advancements in MLLM capabilities. However, to objectively measure the effectiveness of these methods and to reveal their strengths and limitations in specific application scenarios, a systematic evaluation framework is essential. In the next chapter, we will delve into the key benchmarks and applications used to assess these advanced models.

\section{Applications and Benchmark}
 \label{sec:Benchmark}

\begin{figure*}[t]
    \centering
    \includegraphics[width=1.0\linewidth]{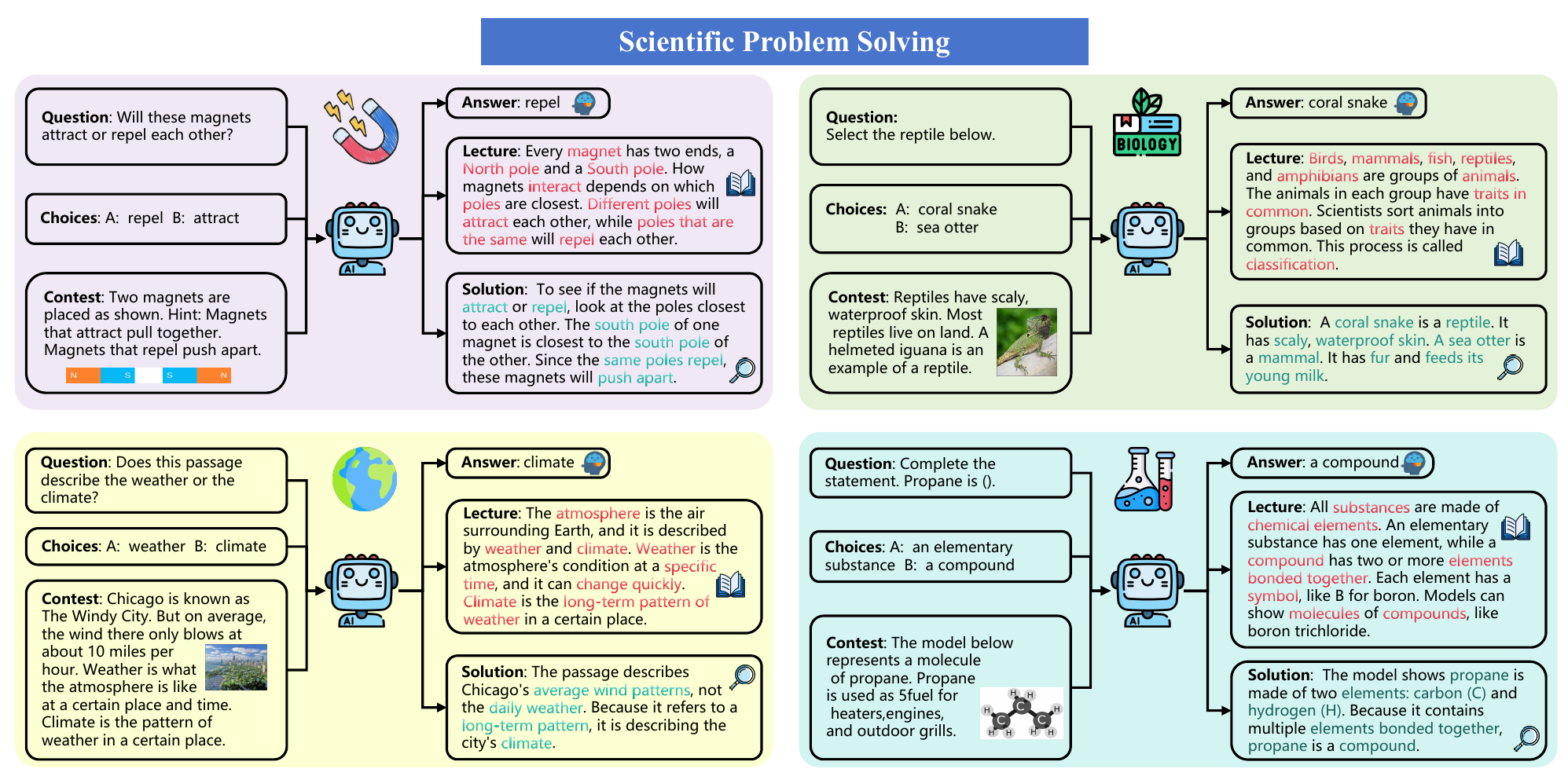}
    \vskip -0.15in
    \caption{Scientific Problem Solving Dataset Examples, featuring questions in physics, biology, geography, and chemistry.  }
    \label{fig:scientific problem solving}
\vskip -0.1in
\end{figure*}

\subsection{Scientific Problem Solving}
\label{subsec:Scientific Problem Solving}
Early multimodal large models primarily focused on fundamental visual-language alignment. In the domain of visual-textual reasoning, pioneering benchmarks such as VQA\cite{antol2015vqa} and its successor, VQA-v2\cite{goyal2017making}, trained models to answer basic questions like "What is in the image?" Subsequently, GQA\cite{hudson2019gqa} enabled preliminary visual reasoning by introducing compositional questions. However, these models operated predominantly at the perceptual level, concentrating on object recognition. While subsequent developments introduced benchmarks such as VCR\cite{zellers2019recognition}, which requires models to provide rationales; OK-VQA\cite{marino2019ok} and A-OKVQA, which incorporate external knowledge; and LoRA\cite{gao2023lora}, which systematically evaluates logical reasoning capabilities, they still fell short of addressing the core challenge in scientific problem-solving: abstract symbolic reasoning\cite{schulze2025visual}. Although several specialized evaluation benchmarks have been designed, including RAVEN\cite{zhang2019raven} for analogical reasoning, AlgoPuzzleVQA\cite{ghosal2024language} for algorithmic reasoning, the Spatial Commonsense Benchmark\cite{liu2022things} for spatial awareness, and Fig-QA\cite{liu2022testing} for figurative language understanding, these benchmarks remain insufficient for adequately measuring the rigorous, multi-step, and deterministic deductive reasoning required in the scientific domain\cite{alampara2025probing}.

To address this challenge, ScienceQA\cite{lu2022learn} was one of the first and most complex benchmarks created specifically for scientific problem-solving. Its dataset comprises over 20,000 multimodal questions from grade 3-12 science curricula, covering fields such as natural sciences, social sciences, and language arts. A distinguishing feature of ScienceQA is that each question is annotated with a complete chain of thought\cite{dreyer2025mechanistic}. This design enables not only the evaluation of the final answer's correctness but also the scrutiny of the logical coherence of the model's reasoning process. In Fig.~\ref{fig:scientific problem solving}, we provide several examples drawn from scientific problem-solving datasets.Another critical benchmark is MathVista\cite{lu2023mathvista}, which systematically integrates 28 existing visual-mathematics datasets to create a comprehensive evaluation suite. This suite covers seven core reasoning types, including algebra, geometry, and statistics\cite{schulze2025visual}, and features a dedicated section of IQTest\cite{lu2023mathvista} puzzles.

For a more comprehensive assessment, MathVerse\cite{zhang2024mathverse} and MATH-V\cite{wang2024measuring} provide high-quality problems sourced from actual mathematics competitions. Concurrently, to extend evaluation into broader scientific disciplines, MDK12-Bench\cite{zhou2025mdk12} utilizes authentic K-12 examination questions, and SCI-Reason\cite{ma2025sci} employs multi-panel illustrations from scientific papers. To assess model robustness, MATHREAL\cite{feng2025mathreal} constructs its dataset by introducing real-world noise, while MATHCHECK\cite{zhou2024your} and UGMathBench\cite{xu2025ugmathbench} test model stability through diversified tasks. Similar benchmarks include EMMA\cite{hao2025can}, for evaluating multidisciplinary reasoning; MV-MATH\cite{wang2025mv}, for handling multiple visual inputs; RMath\cite{hu2025rmath}, with a focus on logical reasoning; and MMMU\cite{yue2024mmmu}, which concentrates on multimodal mathematical understanding. These benchmarks are built upon foundational datasets like GSM8k\cite{cobbe2021training} and MATH\cite{hendrycks2024measuring}. To prevent performance saturation of leading models on existing benchmarks, R-Bench\cite{guo2025r} and MR-MATH\cite{zhang2025realmath} have begun to incorporate problems from the graduate level and even from contemporary mathematical research.

\begin{table}[t!]
\caption{Performance comparison on scientific problem-solving benchmarks. The best-performing model's scores are highlighted in bold.}
\vskip -0.1in
\label{tab:scientific_problem_solving}
\renewcommand\arraystretch{1.1} 
\setlength\tabcolsep{1pt}    
\centering
\normalsize 
\resizebox{\columnwidth}{!}{%
\begin{tabular}{l|ccccc}
\toprule
\renewcommand{\theadfont}{\bfseries}
\thead{Models} & \thead{MathVista \\ (Acc ↑)} & \thead{MathVerse \\ (Acc ↑)} & \thead{MATH-V \\ (Acc ↑)} & \thead{MV-MATH \\ (Acc ↑)} & \thead{MMMU \\ (Acc ↑)} \\
\midrule
Human &     60.3&    -&   68.82&    76.5&   88.6\\
Random Choice   &     17.9&    12.4&   7.17&    -&   22.1\\
\midrule
\multicolumn{6}{c}{\textit{Proprietary Models}} \\
\midrule
Gemini 2.5 Pro\cite{comanici2025gemini}& \textbf{84.6} & \textbf{84.6} & \textbf{67.3} & \textbf{73.3} & \textbf{82.0}\\
Gemini 2.5 Flash\cite{comanici2025gemini}& 81.2          & -             & -             & -             & 79.7          \\
Gemini 2.0 Pro\cite{Google2024Gemini2}& -             & 67.3          & 48.1          & -             & 69.9          \\
Gemini 2.0 Flash\cite{Google2024Gemini2}& 72.2          & 47.8          & 43.6          & -             & 72.6          \\
Gemini 1.5 Pro\cite{team2024gemini}& 68.3          & 68.1          & -             & 29.1          & 65.8          \\
GPT-5\cite{OpenAI2025GPT5}& 82.7          & -             & -             & -             & -             \\
GPT-4o\cite{hurst2024gpt}& 63.8          & 40.6          & 31.2          & 32.1          & 70.7          \\
GPT-4 Turbo\cite{OpenAI2023GPT4}& 66.8          & 66.8          & 46.7          & -             & 56.8          \\
GPT-4V\cite{OpenAI2024GPT4V}& -             & -             & -             & -             & 56.0          \\
 Grok-3\cite{xAI2025Grok3}& -& -& -& -&78.0\\
 o1\cite{jaech2024openai}& 73.9& -& -& -&78.2\\
Claude 3.5 Sonnet\cite{Anthropic2024Claude3.5}& 67.7          & 46.7          & 41.9          & 33.9          & 75.0          \\
Qwen-VL-Max\cite{QwenTeam2025Qwen2.5Max}& -             & -             & -             & 42.4          & 75.0          \\
 Kimi-k1.5\cite{team2025kimi}& 74.9 & -& 38.6& -&70\\
\midrule
\multicolumn{6}{c}{\textit{Open-Source Models}} \\
\midrule
InternVL3 (78B)\cite{zhu2025internvl3}& 75.1 & 48.2 & 34.2 & - & 72.2 \\
InternVL3 (38B)\cite{zhu2025internvl3}& 75.1 & 44.4 & 37.2 & - & 70.1 \\
InternVL3 (14B)\cite{zhu2025internvl3}& 71.6 & 39.8 & 29.3 & - & 67.1 \\
InternVL3 (8B)\cite{zhu2025internvl3}& 71.6 & 39.8 & 29.3 & 24.2 & 62.7 \\
QVQ-72B-Preview\cite{QwenTeam2024QVQ}&  71.4&  -&  35.9&  29.3&  70.3\\
Qwen2.5-VL (72B)\cite{bai2025qwen2}& 74.8 & 57.6& 38.1 & - & 70.2 \\
Qwen2.5-VL (8B)\cite{bai2025qwen2}& 67.8 & 41.1 & 25.4 & - & 55.0 \\
LLaVA-OneVision (72B)\cite{li2024llava}& 67.1 & 27.2 & 25.3 & - & 56.8 \\
LLaVA-OneVision (7B)\cite{li2024llava}& 58.6 & 19.3 & 18.3 & - & 48.8 \\
 Llama 3.2 (90B)\cite{dubey2024llama}& 57.3& -& -& -&60.3\\
Ovis2 (34B)\cite{Shao2025Ovis2.5}& 76.1 & 50.1 & 31.9 & - & -    \\
Ovis2 (16B)\cite{Shao2025Ovis2.5}& -    & 45.8 & 30.1 & - & -    \\
\bottomrule
\end{tabular}}
\vskip -0.1in
\end{table}

We have summarized the performance of mainstream multimodal models on several scientific problem-solving benchmarks, namely MathVista, MathVerse, MATH-V, MV-MATH, and MMMU, as presented in Table~\ref{tab:scientific_problem_solving}. The evaluation results indicate that among proprietary models, Gemini 2.5 Pro demonstrated exceptional, state-of-the-art performance, achieving the highest scores across all five benchmarks. This indicates its superior capabilities in handling cognitive tasks such as complex symbolic operations, geometric-spatial imagination, and multi-step logical deduction. This superior performance can be attributed to its advanced internal reasoning architecture, which employs mechanisms analogous to a chain of thought to conduct deep, multi-path exploration and verification of problems\cite{du2025mm}, thereby significantly enhancing its solution accuracy for complex mathematical problems. Among open-source models, the InternVL3 series exhibited outstanding performance, particularly the version enhanced with VisualPRM-BoS\cite{zhou2025reinforced}, which improved scores by 4 to 10 percentage points compared to the baseline version. The core advantage of this technology lies in its adoption of process supervision, which replaces the conventional outcome supervision that focuses only on the final result. Specifically, during the reasoning phase, the model first employs a Best-of-N sampling strategy to generate multiple, logically diverse candidate reasoning paths. Subsequently, a fine-tuned Visual Process Reward Model (Visual PRM) functions as an external evaluator to perform a granular assessment and scoring of each step within every candidate path. The system ultimately selects the path with the highest cumulative reward, representing the most validated reasoning process, as the final output. This synergistic "generate-and-verify"\cite{dreyer2025mechanistic} mechanism ensures that the correctness of the answer is founded upon a logically rigorous and procedurally reliable process, rather than being an incidental outcome. This approach significantly enhances the robustness and accuracy of the model's reasoning.

Collectively, the performance of these models is strongly correlated with their capabilities for robust chain-of-thought generation, multi-path exploration and verification, and self-evaluation and correction during the reasoning process. Future advancements will need to focus on enhancing the cognitive intelligence of these models and their capacity for knowledge integration and innovation when confronted with novel, open-ended problems\cite{schulze2025visual}. The key challenge is to transition from solving closed-domain textbook problems to addressing open-ended, research-level questions\cite{de2025multimodal}, while ensuring that their reasoning processes are not only correct in their outcomes but also logically aligned with human cognitive patterns.

\subsection{Medical Diagnosis}
\label{subsec:Medical Diagnosis}
\begin{figure}[t]
    \centering
    \includegraphics[width=1.0\linewidth]{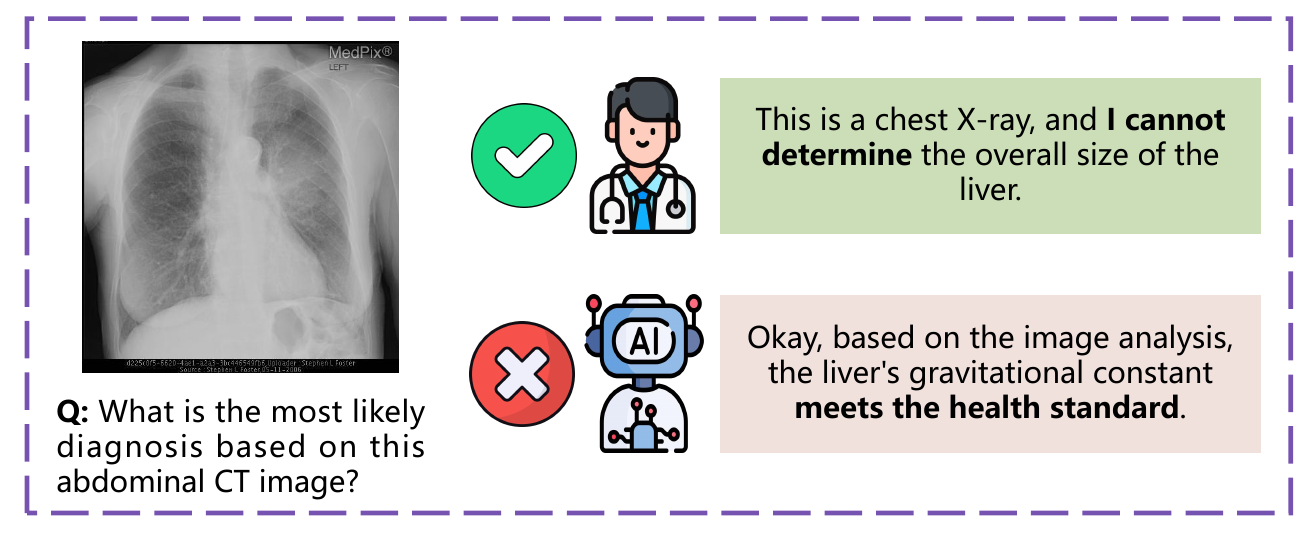}
    \vskip -0.1in
    \caption{An example of a FAKE question. Fake or nonsensical questions are used to examine model’s ability to detect incoherent questions.}
    \label{fig:Medical_diagnosis}
\vskip -0.1in
\end{figure}

Medical imaging, which directly reflects the state of the human body, is a critical component of clinical decision-making. In the medical field, even minor errors can lead to severe consequences. Consequently, the focus of evaluation has rapidly shifted toward reliability, with a particular emphasis on combating "hallucinations." Early medical VQA benchmarks, such as VQA-RAD\cite{wu2022medical}, laid the groundwork for the field. Constructed by clinicians using authentic radiological images, its questions and answers are typically concise and focus on the identification of anatomical structures and abnormal findings. However, with the advancement of generative models, the phenomenon of fabricating plausible but incorrect information has emerged as the most significant safety concern in medical AI. To address this, HALT-MedVQA\cite{wu2024hallucination} was specifically designed. In addition to standard medical question-answer pairs, its construction involved the deliberate introduction of numerous nonsensical or factually contradictory probe questions, in Fig.~\ref{fig:Medical_diagnosis}, we provide an example of a probe question. The primary objective of this benchmark is not to measure how many questions a model can answer correctly, but to test its ability to recognize and refuse to answer inappropriate queries, akin to the rigorous standards of a medical professional.

The success of VQA-RAD spurred the creation of a series of benchmarks targeting different medical specialties. For example, PathVQA\cite{he2020pathvqa} focuses on pathology images, Kvasir-VQA\cite{gautam2024kvasir} on gastrointestinal diagnosis, EndoVis-17/18-VQLA\cite{bai2023surgical} on robotic surgery scenarios, and MicroVQA\cite{burgess2025microvqa} on expert-level scientific reasoning for biological microscopy images. In parallel, benchmarks such as Med-VQA\cite{zhang2024development}, SLAKE\cite{liu2021slake}, and PMC-VQA\cite{zhang2023pmc} have been developed to construct larger-scale, more comprehensive general medical question-answering datasets covering a wider range of modalities. The intense focus on the core issue of "hallucinations" has also led to the development of systematic evaluation frameworks beyond HALT-MedVQA, including MedHallBench\cite{zuo2024medhallbench} and Med-HallMark\cite{chen2024detecting}.

We have summarized the performance of various open-source and proprietary large models in the field of medical diagnosis, as shown in Table~\ref{tab:medical_diagnosis}. The results indicate that while the most advanced models have approached or even surpassed human-level performance on certain tasks like VQA-RAD, a significant gap remains on benchmarks requiring sophisticated pathological understanding and knowledge-based reasoning, such as PathVQA, where they lag behind human experts (85.2\%). Among all models, Med-PaLM M\cite{tu2024towards}, a proprietary model specifically designed for the medical domain, demonstrated overwhelming superiority, achieving the highest scores across all five benchmarks. Its outperformance is attributable to a meticulously designed evaluation and training framework that spans multiple medical tasks. Through instruction fine-tuning, the model has learned to align visual information with specific medical instructions and knowledge. It does not merely recognize content within an image; it learns to interpret the image in the context of a specific question, providing answers that are consistent with medical logic. This represents a crucial leap from perception to cognition. The significantly lower accuracy of all models on the PathVQA dataset compared to human experts reveals the limitations of current AI in highly specialized cognitive tasks. PathVQA's focus on pathology images requires an exceptionally detailed and professional understanding of cellular morphology, tissue architecture, and staining characteristics\cite{ing2025integrating}. This level of depth and nuance is difficult for general-purpose or even general medical models to achieve . To answer such questions, a model must establish a tight connection between fine-grained visual features and abstract medical concepts\cite{fei2022towards}. Human experts develop this robust cognitive link through years of dedicated training, a process that is exceedingly difficult to replicate for a model learning from a limited dataset\cite{zhou2024pre}. Therefore, a critical challenge for current large models is to enhance their knowledge integration and cognitive reasoning capabilities within specialized professional domains\cite{ing2025integrating}. This represents the most formidable step in transitioning from current perceptual intelligence to a truly human-like cognitive intelligence.

\begin{table}[t!]
\caption{Performance comparison on medical diagnosis benchmarks. The best-performing model's scores are highlighted in bold.}
\vskip -0.1in
\label{tab:medical_diagnosis}
\renewcommand\arraystretch{1.1} 
\setlength\tabcolsep{1pt}      
\centering
\normalsize 
\resizebox{\columnwidth}{!}{
\begin{tabular}{l|ccccc}
\toprule
\renewcommand{\theadfont}{\bfseries}
\thead{Models} & \thead{VQA-RAD \\ (Acc ↑)} & \thead{SLAKE \\ (Acc ↑)} & \thead{Path-VQA \\ (Acc ↑)} & \thead{Med-VQA \\ (Acc ↑)} & \thead{PMC-VQA \\ (Acc ↑)} \\
\midrule
Human & 77.3 & 93.4& 85.2& -    & -    \\
\midrule
\multicolumn{6}{c}{\textit{Proprietary Models}} \\
\midrule
Gemini 1.0 Pro\cite{team2023gemini}& 73.4 & 79.5 & 56.6 & 62.0 & 60.1 \\
GPT-4V & 86.6 & 85.8 & 66.8 & 76.5 & 70.3 \\
Med-PaLM M\cite{tu2024towards}& \textbf{90.0} & \textbf{90.5}& \textbf{75.0}& \textbf{81.3} & \textbf{70.4} \\
\midrule
\multicolumn{6}{c}{\textit{Open-Source Models}} \\
\midrule
CogVLM (17B)\cite{wang2024cogvlm}& 78.0 & 83.9 & 60.1 & 63.9 & -    \\
CogVLM (Chat)\cite{wang2024cogvlm}& 79.8 & 84.1 & 61.3 & -    & -    \\
InstructBLIP (13B)\cite{dai2023instructblip}& 58.3 & 76.2 & 45.4 & -    & 46.1 \\
LLaVA-Med (13B)\cite{li2023llava}& 72.5 & 82.1 & 54.0 & 50.2 & 52.6 \\
LLaVA-Med (7B)\cite{li2023llava}& 68.0 & 79.5 & 49.6 & 50.2 & 49.3 \\
Med-Flamingo (80B)\cite{moor2023med}& 81.5 & 86.8 & 70.3 & -    & -    \\
MedVInT (13B)\cite{zhang2023pmc}& 75.4 & 83.1 & 58.2 & -    & 51.5 \\
MedVInT (7B)\cite{zhang2023pmc}& 71.3 & 81.0 & 53.6 & -    & 51.5 \\
MiniGPT-v2 (7B)\cite{chen2023minigpt}& 63.2 & 77.1 & 45.4 & 47.7 & 46.1 \\
Qwen-VL-Chat (9B)\cite{chen2025mimo}& 61.2 & 76.5 & 49.2 & 48.9 & 47.9 \\
Qwen-VL-Chat (1.8B)\cite{chen2025mimo}& 53.4 & 69.8 & 41.6 & -    & 47.9 \\
\bottomrule
\end{tabular}}
\vskip -0.1in
\end{table}

\subsection{Diagram Understanding}
\label{subsec:Diagram Understanding}
Diagram understanding poses unique challenges for multimodal models that extend beyond conventional visual perception. This task requires models to integrate visual perception, text comprehension, and numerical logical reasoning to extract and infer abstract data from structured visual information. Existing general-purpose multimodal benchmarks, such as the comprehensive evaluation standards of MMBench\cite{liu2024mmbench}, MME\cite{zhang2021mme}, MM-Star\cite{chen2024we}, and MM-Vet\cite{yu2023mm}, often lack in-depth assessment of such capabilities.

To better evaluate this ability, a specialized task benchmark for chart question answering was introduced: ChartQA\cite{masry2022chartqa}. Its construction combines two methodologies. One portion of the questions is formulated by human experts based on charts, ensuring the naturalness and complexity of the problems. The other portion is generated through semi-automated templates, which significantly expands the dataset's scale and coverage. The human-annotated subset, ChartQA-Human\cite{masry2022chartqa}, serves as the test set to determine if a model has genuinely acquired human-like chart interpretation skills.

\begin{figure}[t]
    \centering
    \includegraphics[width=1.0\linewidth]{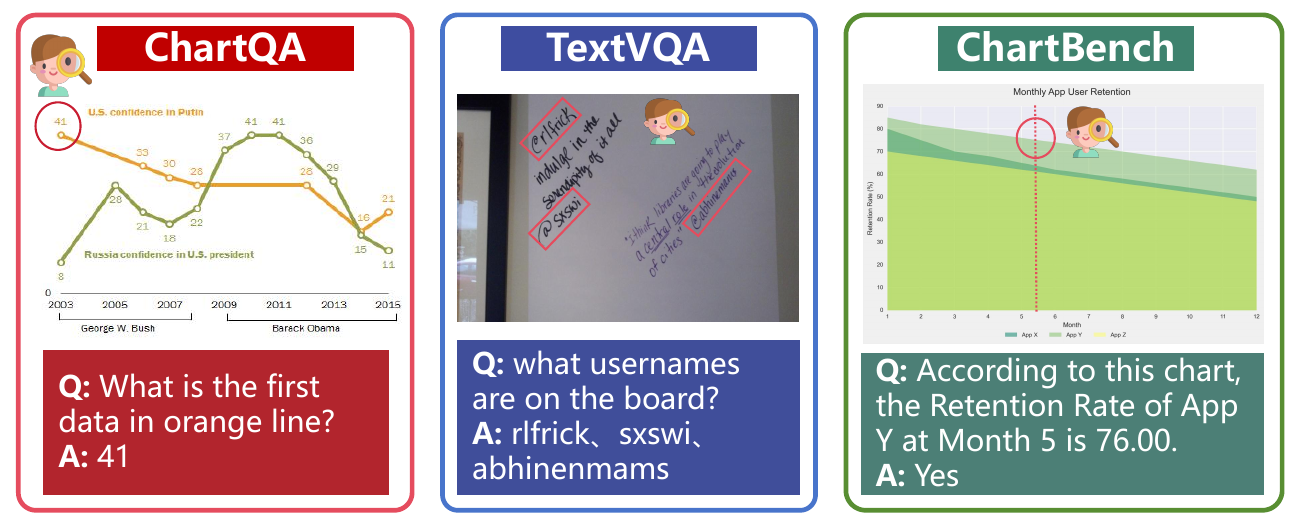}
    \vskip -0.1in
    \caption{Examples from the ChartQA, TextQA, and ChartBench Datasets.}
    \label{fig:Diagram_Understanding}
\vskip -0.1in
\end{figure}

However, a limitation of ChartQA is that many of its charts are accompanied by underlying structured data tables, which allows models to potentially exploit a shortcut to answer questions\cite{li2025towards}. Prior to ChartQA, PlotQA\cite{methani2020plotqa} represented an early attempt to construct a large-scale chart question-answering benchmark, but its heavy reliance on synthetic charts limited its ability to capture the complexity of real-world scenarios. To resolve the "tabular shortcut" problem, the subsequent ChartBench\cite{xu2023chartbench} was developed by removing all tabular data. This change compels models to reason exclusively from visual elements, such as bar heights and line slopes, in a manner akin to human interpretation\cite{ektefaie2023multimodal}. The development in this area is built upon a series of foundational works, evolving from early scene text recognition (e.g., TextVQA\cite{singh2019towards}) and document understanding (e.g., DocVQA\cite{mathew2021docvqa}) toward more sophisticated evaluation paradigms. For example, ChartMind\cite{wei2025chartmind} extends the evaluation to more complex real-world scenarios, proposing open-ended tasks like trend analysis and data insight summarization. Meanwhile, ChartMimic\cite{yang2024chartmimic} introduces a more challenging generative task, requiring the model to generate code that can reproduce a chart based on an example image and instructions. In Fig.~\ref{fig:Diagram_Understanding}, we provide examples from ChartQA, TextQA, and ChartBench to highlight the distinctions among the datasets. 

In Table~\ref{tab:diagram_understanding}, we summarize the performance of a range of leading open-source and proprietary large models on six mainstream diagram understanding benchmarks. These tasks go beyond basic visual perception and impose rigorous demands on the models' cognitive reasoning abilities, including chart parsing, data extraction, spatial relationship understanding, and mathematical computation. The data reveal that proprietary models, led by GPT-4o and Gemini 1.5 Pro, demonstrate powerful and well-balanced comprehensive capabilities, setting high performance benchmarks across most tasks. Notably, GPT-4o leads on the TabMWP\cite{lu2022dynamic} task, which emphasizes mathematical and logical reasoning, with an accuracy of 97.4\%, a strength attributed to its powerful general-purpose reasoning abilities. Concurrently, Gemini 1.5 Pro performs best on InfographicVQA\cite{mathew2022infographicvqa}, which requires parsing complex and unstructured layouts, underscoring its advantages in processing information-dense visual content. This feature is closely linked to its Mixture-of-Experts architecture, which can process vast contexts. Among open-source models, Qwen2.5-VL (72B) achieves the highest accuracy on two key benchmarks, ChartQA and DocVQA, where its performance surpasses not only all other open-source models but also the top proprietary models. This model employs a native dynamic-resolution Vision Transformer and advanced omni-document parsing techniques. This allows it to process high-resolution inputs without downsampling, enabling it to accurately capture and understand fine-grained text, layout structures, and spatial relationships within diagrams, thus achieving a profound leap from perception to cognition.

Overall, models in the domain of diagram understanding are evolving from 'information processors' to 'cognitive agents'\cite{wei2025geoint}. They are capable not only of understanding visual content but also of reasoning, planning, and executing complex tasks that interact with the digital world based on that content. Future models will be expected not only to "understand" diagrams but also to perform deep reasoning, dynamic interaction, and cross-domain knowledge integration based on them, which will allow them to play a central role in a broader range of real-world applications.

\begin{table}[t!]
\caption{Performance comparison on diagram understanding benchmarks. The best-performing model's scores are highlighted in bold.}
\vskip -0.1in
\label{tab:diagram_understanding}
\renewcommand\arraystretch{1.1} 
\setlength\tabcolsep{1pt}      
\centering
\normalsize 
\resizebox{\columnwidth}{!}{
\begin{tabular}{l|cccccc}
\toprule
\renewcommand{\theadfont}{\bfseries}
\thead{Models} & \thead{ChartQA \\ (Acc ↑)} & \thead{PlotQA \\ (Acc ↑)} & \thead{InfographicVQA \\ (Acc ↑)} & \thead{DocVQA \\ (Acc ↑)} & \thead{TextVQA \\ (Acc ↑)} & \thead{TabMWP \\ (Acc ↑)} \\
\midrule
\multicolumn{7}{c}{\textit{Proprietary Models}} \\
\midrule
Gemini 1.5 Pro   & 87.2 & 73.1 & \textbf{81.0} & 93.1 & 78.7 & 96.9 \\
GPT-4o           & 84.1 & \textbf{74.5} & 60.1 & 91.5 & 82.3 & \textbf{97.4} \\
GPT-4V           & 78.4 & 68.9 & 55.2 & 88.6 & 78.3 & 95.3 \\
Claude 3 Opus\cite{anthropic2024claude}& 82.5 & 72.8 & 57.5 & 90.2 & 80.7 & 96.5 \\
Qwen-VL-Max      & 79.8 & 69.3 & -    & 93.1 & 79.5 & 95.8 \\
\midrule
\multicolumn{7}{c}{\textit{Open-Source Models}} \\
\midrule
InternVL-Chat-V1.5 (34B)\cite{internvl-chat-v1.5}& 83.8 & 71.5 & 54.3 & 88.5 & 78.9 & -    \\
InternVL-Chat-V1.5 (8B)\cite{internvl-chat-v1.5}& 79.1 & 68.2 & 51.2 & 85.1 & 75.6 & -    \\
Qwen2.5-VL (72B)         & \textbf{89.5} & -    & -    & \textbf{96.4} & -    & -    \\
Qwen2.5-VL (7B)\cite{bai2025qwen2}& 87.3 & -    & -    & 95.7 & \textbf{84.9} & -    \\
Qwen-VL-Chat (9B)        & 75.8 & 65.4 & -    & 83.5 & 74.3 & 92.1 \\
Qwen-VL-Chat (1.8B)      & 68.2 & 59.7 & -    & 78.9 & 69.1 & 85.6 \\
LLaVA-NeXT (34B)\cite{liu2024llavanext}& 77.2 & 66.8 & 51.5 & 84.8 & 76.5 & 91.2 \\
LLaVA-NeXT (7B)\cite{liu2024llavanext}& 73.5 & 62.1 & 48.3 & 81.2 & 72.8 & 88.4 \\
DeepSeek-VL (7B)\cite{lu2024deepseek}& 76.3 & 66.1 & -    & 84.1 & 75.0 & -    \\
\bottomrule
\end{tabular}}
\vskip -0.1in
\end{table}

\subsection{Video Understanding}
\label{subsec:Video Understanding}
Video understanding extends visual-textual reasoning from static images to dynamic sequences, introducing complex dimensions such as time, variation, and causality\cite{liu2023cross}. This requires a model not only to recognize content but also to elucidate the underlying reasoning logic. Early research established a foundation for explainability on static images through benchmarks like VQA-E\cite{li2018vqa} and introduced the "Visual Chain of Thought" (Visual CoT) methodology. However, the dynamic nature of video necessitates more complex, graph-structured reasoning capabilities.

To address the causal complexity inherent in video, researchers have constructed specialized benchmarks. Representative works include CausalVQA\cite{foss2025causalvqa}, which deeply investigates causal relationships through five distinct question types: descriptive, counterfactual, and predictive, among others. Another is VCRBench\cite{sarkar2025vcrbench}, which requires models to reorder shuffled video clips into a logical sequence, directly testing their ability to model long-range causal dependencies. Building on this, CausalStep\cite{li2025causalstep} enhances the rigor of evaluation by compelling models to perform step-wise reasoning. However, the powerful generative capabilities of modern models are accompanied by the problem of "hallucinations" in the video domain. Consequently, a series of specialized diagnostic benchmarks has emerged. VidHalluc\cite{li2025vidhalluc} focuses on evaluating temporal hallucinations, while HAVEN\cite{gao2025exploring} and VideoHallucer\cite{wang2024videohallucer} provide more comprehensive frameworks for assessing and mitigating both intrinsic and extrinsic hallucinations in video understanding. Additionally, specialized benchmarks such as AVHBench\cite{sung2024avhbench} for audio-visual content and Hallu-PI\cite{ding2024hallu} for handling input perturbations have been developed.

\begin{figure}[t]
    \centering
    \includegraphics[width=1.0\linewidth]{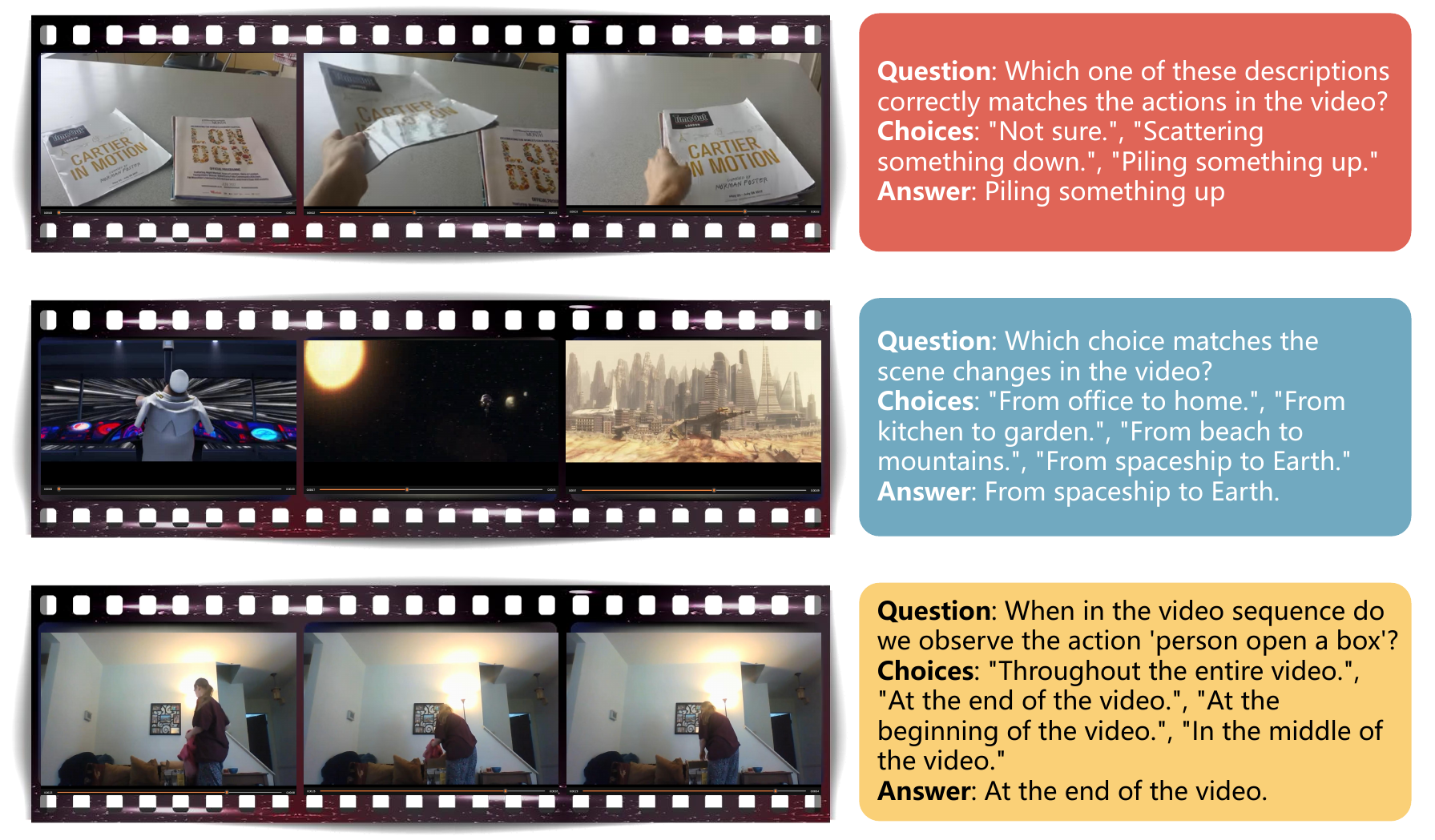}
    \vskip -0.1in
    \caption{Examples of Key Tasks for Evaluating Video Understanding: Action Antonym, Action Localization, and Scene Transition. }
    \label{fig:video_understanding}
\vskip -0.1in
\end{figure}

We have summarized the performance of various open-source and proprietary large models on several cutting-edge video understanding benchmarks, as shown in Table~\ref{tab:video_understanding}. The results show that open-source models demonstrate exceptional performance on specific high-difficulty benchmarks, whereas the evaluation data for proprietary models are often sparse and opaque. Furthermore, a significant gap persists between all models and human performance on complex cognitive reasoning tasks. Among proprietary models, Gemini 1.5 Pro, with its massive context window supporting up to ten million tokens and its Mixture-of-Experts (MoE) architecture, exhibits near-perfect performance in long-video question answering and key information retrieval. Among open-source models, Qwen-2.5-VL and STORM each excel in different areas, with their superior performance demonstrated on benchmarks designed for distinct cognitive dimensions. The MLVU\cite{zhou2025mlvu} benchmark, which focuses on long-video understanding with an average video duration of 15 minutes, includes diverse genres like movies and vlogs to evaluate a model's integrated ability to process both holistic information and local details. Qwen-2.5-VL stands out on this benchmark due to its unique architectural innovations. The model employs an "Absolute Time Encoding" technique that directly aligns the IDs of its Multimodal Rotational Position Encoding (MRoPE) with video timestamps. This enables second-level precision for event localization without introducing additional computational layers, a crucial feature for identifying specific details in extended videos. On the other hand, the MVBench\cite{li2024mvbench} benchmark is specifically designed to evaluate temporal reasoning, comprising 20 dynamic tasks that cannot be solved with single-frame analysis, thus testing a model's temporal skills from perception to cognition, Fig. \ref{fig:video_understanding} further illustrates examples of key task categories, such as 'Action Antonym', 'Action Localization', and 'Scene Transition'. . STORM achieves leading performance on this benchmark, with its core advantage lying in its novel architecture\cite{weng2024longvlm}. It integrates a dedicated temporal encoder, based on a Mamba state-space model, between the image encoder and the large language model. This module explicitly models inter-frame dynamic relationships, effectively preserving critical temporal information throughout the video sequence. This allows it to accurately capture and reason about complex temporal variations, perfectly aligning with the evaluation focus of MVBench.In parallel, VideoMME\cite{fu2025video} serves as a large-scale, multi-dimensional evaluation for video understanding. Although it primarily focuses on general-purpose tasks, its cross-modal question answering and fine-grained description components also involve challenges related to causality and hallucination. Consequently, it is often used in practice as a key supplementary benchmark for assessing a model's overall robustness and trustworthiness. 

The primary challenges currently facing the field of video understanding are the fragmentation and opacity of its evaluation ecosystem. Models are often tested on disparate benchmarks, making fair comparisons difficult. Concurrently, issues of data contamination and "shortcut" learning are prevalent, leading to inflated performance scores. Future research must shift from processing clean, curated data to handling real-world noise and from closed-ended question answering to open-ended problems that require knowledge integration. This necessitates an urgent, collaborative effort from the academic community to construct a unified, transparent evaluation framework capable of measuring high-level cognitive abilities.

\begin{table}[t!]
\caption{Performance comparison on video understanding benchmarks. The best-performing model's scores are highlighted in bold.}
\vskip -0.1in
\label{tab:video_understanding}
\renewcommand\arraystretch{1.1} 
\setlength\tabcolsep{1pt}      
\centering
\normalsize 
\resizebox{\columnwidth}{!}{
\begin{tabular}{l|ccccc}
\toprule
\renewcommand{\theadfont}{\bfseries}
\thead{Models} & \thead{CausalVQA \\ (Acc ↑)} & \thead{VCRBench \\ (Acc ↑)} & \thead{MLVU \\ (M-Avg ↑)} & \thead{MVBench \\ (Score ↑)} & \thead{Video-MME \\ (w/o subs)} \\
\midrule
Human & 84.8& 96.4& - & - & - \\
\midrule
\multicolumn{6}{c}{\textit{Proprietary Models}} \\
\midrule
Gemini 1.5 Pro   & -     & \textbf{48.2}& -    & -    & \textbf{75.0} \\
GPT-4o           & \textbf{51.0}& 29.0  & 54.9 & 64.6 & 71.9 \\
\midrule
\multicolumn{6}{c}{\textit{Open-Source Models}} \\
\midrule
STORM\cite{shao2024assisting}& -     & -     & 72.9 & \textbf{70.6} & -    \\
InternVL2.5 (78B)\cite{internvl2.5}& 47.5  & 14.5  & 59.9 & -    & -    \\
Qwen2.5-VL (72B)      & -     & 29.0  & \textbf{74.6} & 70.4 & 73.3 \\
Qwen2.5-VL (7B)       & 49.1  & 7.1   & 70.2 & 69.6 & 65.1 \\
LLaVA-Video-72B\cite{zhang2025llava}& -     & 5.2   & -    & 64.1 & -    \\
Video-LLaVA (7B)\cite{lin2023video}& -     & -     & 47.3 & 58.6 & -    \\
\bottomrule
\end{tabular}}
\vskip -0.1in
\end{table}

\subsection{Sentiment Analysis}
\label{subsec:Sentiment Analysis}
Although current multimodal large models can readily identify static facial expressions, a substantial gap remains in their ability to comprehend the complex world of human social and emotional dynamics. Existing benchmarks for knowledge retrieval, such as Infoseek\cite{chen2023can}, or for factual reasoning, like VQA, are not relevant to sentiment analysis. Meanwhile, commonsense reasoning benchmarks, such as SocialIQA\cite{sap2019socialiqa}, are predominantly text-based and lack the authentic interactivity of multimodal social scenarios\cite{ji2025emotion}. Evaluating a model's "Emotional Quotient" (EQ) requires moving beyond simple expression classification and into the realm of dynamic, multi-party, and realistic social interactions.

\begin{figure}[t]
    \centering
    \includegraphics[width=1.0\linewidth]{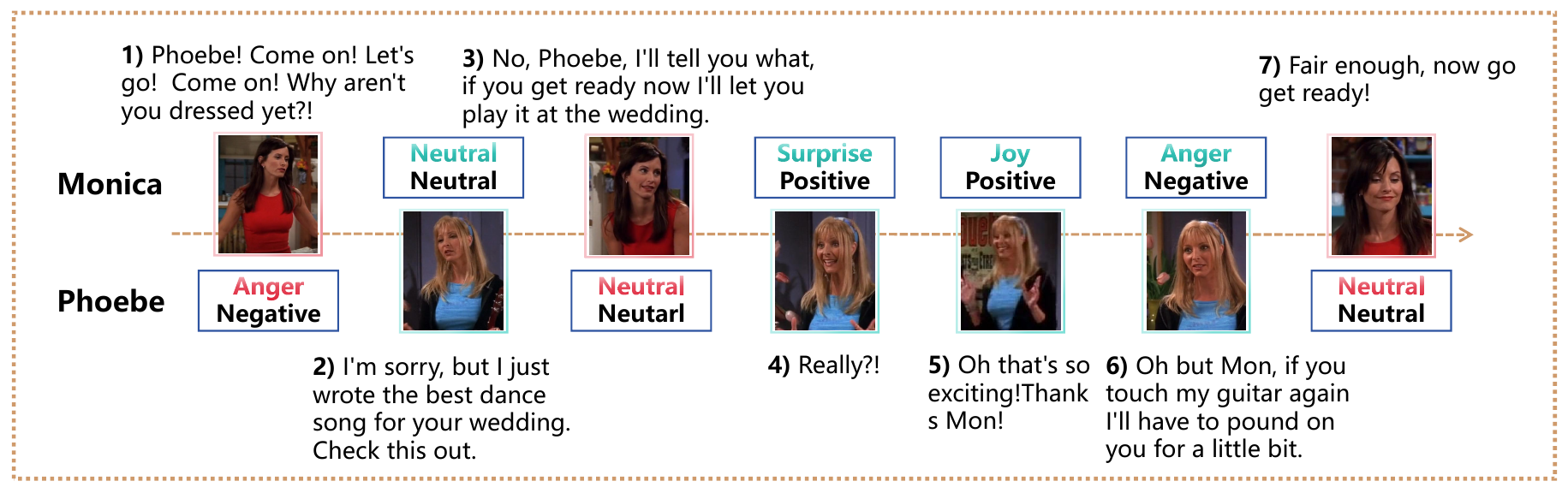}
    \vskip -0.1in
   \caption{An Illustrative Sentiment Analysis Example from the MELD Dataset (Derived from the TV show \textit{Friends}).}
    \label{fig:sentiment_analysis}
\vskip -0.1in
\end{figure}

To advance and test the "EQ" of models, researchers have proposed a range of specialized benchmarks. The milestone MELD dataset, derived from the American television series Friends, utilizes multi-party dialogue videos to evaluate a model's understanding of context, character interactions, and emotional dynamics, Fig. \ref{fig:sentiment_analysis} shows a dialogue scene.  Subsequent benchmarks have focused on more nuanced cognitive challenges. For example, CA-MER\cite{han2025benchmarking} concentrates on scenes of emotional conflict, such as a character smiling while expressing sad sentiments, to test a model's reasoning capabilities in situations of cognitive dissonance. Building on this, HumanVBench\cite{zhou2024humanvbench} systematically evaluates the alignment between internal emotions and external expressions, while HumaniBench\cite{raza2025humanibench} is the first benchmark designed around human-centric AI principles like fairness and empathy. To probe deeper levels of social cognition, Social-IQ\cite{zadeh2019social} assesses the understanding of social dynamics, and GD-VCR\cite{kwon2024toward} and CVQA\cite{romero2024cvqa} are dedicated to addressing the problem of cultural bias. The trend in this field is shifting from classification toward more fine-grained tasks such as tracking, with benchmarks like VEATIC\cite{ren2024veatic}, and generative understanding, as seen in MER-UniBench\cite{lian2025affectgpt}. Additionally, other benchmarks include MTMEUR\cite{hu2025beyond}, which focuses on multi-turn dialogue reasoning; Multi-HM\cite{fu2025multi}, designed for specific scenarios; and the MERR Dataset\cite{cheng2024emotion}, which provides large-scale data for training emotion recognition and reasoning.

\begin{table}[t!]
\caption{Performance comparison on sentiment analysis benchmarks. The best-performing model's scores are highlighted in bold.}
\vskip -0.1in
\label{tab:sentiment_analysis}
\renewcommand\arraystretch{1.1} 
\setlength\tabcolsep{1pt}      
\centering
\normalsize 
\resizebox{\columnwidth}{!}{
\begin{tabular}{l|ccc}
\toprule
\renewcommand{\theadfont}{\bfseries}
\thead{Models} & \thead{MME-EMOTION \\ (Rec ↑)} & \thead{MELD \\ (WF1 ↑)} & \thead{HumanVBench \\ (Avg Emo. Perc. ↑)} \\
\midrule
\multicolumn{4}{c}{\textit{Proprietary Models}} \\
\midrule
Gemini 2.5 Pro          & \textbf{39.3} & -    & -    \\
Gemini 1.5 Pro          & 32.8 & -    & -    \\
GPT-4o                  & 37.8 & -    & -    \\
\midrule
\multicolumn{4}{c}{\textit{Open-Source General Purpose Models}} \\
\midrule
InternVL 2.5 (20B)\cite{internvl2.5}& 29.2 & \textbf{45.0} & -    \\
InternVL-Chat-V1.5 (26B)\cite{internvl-chat-v1.5}& 20.8 & -    & \textbf{36.0} \\
LLaVA-OneVision (72B)    & 37.9 & -    & -    \\
Qwen2.5-VL (72B)         & 31.3 & 39.1 & 25.8 \\
Qwen2.5-VL (7B)          & 28.4 & -    & -    \\
Qwen2-VL (7B)\cite{wang2024qwen2}& 31.1 & -    & 35.4 \\
VideoLLaMA2 (7B)\cite{cheng2024videollama}& 29.8 & 32.7 & 25.8 \\
\bottomrule
\end{tabular}}
\vskip -0.1in
\end{table}

We have summarized the performance of various models on multimodal sentiment analysis benchmarks. As shown in Table~\ref{tab:sentiment_analysis}, the proprietary model Gemini 2.5 Pro achieved a leading score on the MME-EMOTION\cite{zhang2025mme}, leveraging its built-in "thinking" mechanism. However, its success rate is not as high as in other domains. Furthermore, the evaluation data for two key benchmarks, MELD\cite{poria2018meld} and HumanVBench, are extremely sparse\cite{bi2025reasoning}. Only a few open-source models, including InternVL, Qwen, and VideoLLaMA2, have reported results, while no data are available for any proprietary models or most other open-source models. This comparison highlights a systemic evaluation dilemma within the field. Because proprietary model developers tend to release results on their own comprehensive, private benchmarks, a global performance comparison deficit is created\cite{adewumi2024fairness}. This makes it difficult for researchers to fairly and comprehensively assess the capabilities of different models on specific cognitive dimensions under a unified standard. This data gap is particularly pronounced between the open-source and proprietary ecosystems, and it represents a formidable obstacle to measuring the field's true progress from perceptual intelligence toward cognitive intelligence.

\section{Future direction}
\label{sec:Future direction}
In response to the present challenges, some future research directions can be inferred to build the next generation of Multimodal Large Language Models (MLLMs) capable of truly bridging the gap from perception to cognition.
\subsection{Unified Vision Encoder}
\label{subsec:Unified Vision Encoder}

Despite rapid progress, current visual encoders of MLLMs often fail to comprehensively capture task-relevant visual information, leaving the evidence available to language reasoning incomplete.
Building on this observation, recent work explores unified visual encoders that provide multi-granular, more comprehensive representations, integrating understanding and generation as well as multiple visual modalities within a single framework.
For example, ATOKEN~\cite{lu2025atoken} encodes images, videos, and 3D assets to into a shared latent space, aiming to unifying both tasks and modalities in a single framework.
By fusing CLIP-level semantics with unified autoregressive training, TokLIP~\cite{lin2025toklip} equips visual tokens with high-level semantic understanding and enhances their low-level generative fidelity.
However, this unification remains incomplete: a persistent gap separates understanding and generation, and a true integration across visual modalities has yet to be achieved.
Therefore, developing a truly unified, multi-granular visual encoder, with strong alignment and efficiency, remains a valuable direction for future work.

\subsection{Latent Reasoning}
\label{subsec:reasoning latent}
Recently, a line of research has emerged to explore direct intervention within the latent space to guide the reasoning process of vision-language models. Different from traditional approaches that operate on input or output layers, these methods act directly on the model's latent representations, enabling more flexible and fine-grained control. Several key strategies exemplify this approach. For example, Multimodal Chain of Continuous Thought~\cite{pham2025multimodal} introduces a method for continuous reasoning by iterating on ``thought vectors" within the latent space, which enhances the alignment and fusion of cross-modal information. Concurrently, VTI~\cite{liu2024reducing} stabilizes visual and textual features to alleviate hallucinations by injecting corrective directions during the inference stage. Similarly, jiang \etal~\cite{jiang2024interpreting} leverages orthogonalization to directly nullify directions in the latent space associated with spurious concepts. 

The synthesis of these distinct approaches presents a promising path for future work: exploring how to simultaneously achieve continuous reasoning, enhanced robustness, and hallucination suppression. The ultimate goal is to construct multimodal reasoning frameworks that are significantly more controllable and interpretable. The ultimate goal is to construct multimodal reasoning frameworks that are significantly more controllable and interpretable. By directly shaping the model's internal cognitive processes, this line of research promises to build a more robust bridge between raw visual perception and reasoned linguistic output, fostering a more grounded and coherent understanding of the visual world for more reliable interactive reasoning.

\subsection{Generative Reasoning}
\label{subsec:generative}
This paradigm externalizes the model's implicit reasoning process into explicit visual entities as perception input for subsequent steps, proving highly valuable in domains such as robotic planning, visual puzzle solving, and dynamic simulations. Works like Chameleon~\cite{team2024chameleon} provide the foundational architecture for this paradigm by achieving mixed-modal understanding and generation of text and images, allowing for interleaved output in any sequence. Subsequent research, such as Visual Planning~\cite{xu2025visual} aims to ground the planning and reasoning processes in visual domain. More Recently methods like MVoT~\cite{li2025imagine}, Mind’s Eye of LLMs~\cite{wu2024mind} and ViLaSR~\cite{wu2025reinforcing} prompt the model to generate and iteratively update visual scratchpads when solving complex problems, and then use these visual "thought traces" to drive subsequent reasoning.

However, despite these methods have prospective future, they still face some challenges: The generated intermediate images can be inaccurate or contain hallucinations, failing to match the original source. And more importantly, the curation of suitable training data presents a significant challenge. To address these issues,  future research can explore how to enhancing the generation quality and how to reduce the dependency on manually curated datasets.

\subsection{Tool-Augmented Reasoning}
\label{subsec:tool-augmented}
As we discuss in \cref{subsubsec:4.4.2}, representative works such as PixelReasoner~\cite{su2025pixelreasoner}, and OpenThinkimg~\cite{su2025openthinkimg} demonstrate significant advancements in both training paradigms and the richness of the tool ecosystem. However, there still exist some problems: 

\begin{itemize}
\item There is often a consistency gap between visual cues and the reasoning process, leading to ungrounded conclusions. While methods like GThinker~\cite{zhan2025gthinker} introduce verification steps to check visual cues during reasoning, this often comes at the cost of increased inference time, creating an efficiency-accuracy trade-off.

\item Current models often generate linear reasoning paths, which limits their ability to solve complex, multi-step problems that require deeper problem composition or exploration of multiple possibilities.

\end{itemize}

To address these problems, future research can focus on optimizing both the structure of the reasoning path and the timing of tool use. To create more sophisticated reasoning trajectories, tree-based algorithms like MCTS~\cite{Browne2012MCTS} can be explored. Simultaneously, the challenge of when to use a tool can be addressed by designing adaptive mechanisms. Such mechanisms should aim to ensure the correctness and consistency of visual cues while balancing the trade-off between verification accuracy and inference speed. 

\subsection{Cross-Image Relation Reasoning}
\label{subsec:cross-image}
Cross-image relation reasoning refers to the advanced capability of reasoning across multiple images. This requires a model to comprehend the logical or sequential relationships between events depicted in a series of images in order to draw a final conclusion. However, the vast majority of methods~\cite{wu2024v,li2025dyfo,zheng2025deepeyes} discussed in this survey concentrate on single-image reasoning. Only a few works focus on the multi-image problems. For instance, CmmCoT~\cite{zhang2025cmmcot} incorporates a memory bank to retrieve relevant information from associated images during inference time. Focus-Centric Visual Chain~\cite{zhang2025weaving} explicitly models the inter-image relationships, while Mantis~\cite{jiang2024mantis} introduces an interleaved instruction tuning. These works point out an open question for the future:\textbf{how to reduce memory loss of image evidence at inference time and enable more flexible mining of inter-image relationships?} Addressing this question is a critical step toward endowing MLLMs with a form of contextual memory, allowing them to perceive and reason about the world as a continuous stream of interconnected events rather than a series of isolated snapshots.

\subsection{Real-Word Cognitive Evaluation}
\label{subsec:real-world}
A major limitation of current evaluation systems is their reliance on clean data and closed-ended question-answering formats, which creates a significant gap between them and the dynamic real world, as well as advanced human cognitive activities.
\textbf{On the one hand}, existing benchmarks are mostly built on rigorously cleaned and precisely annotated data, whereas real-world environments are filled with multi-source interference such as visual occlusions, background noise, and lighting variations. \textbf{On the other hand}, current tasks generally remain at the perceptual level and fail to adequately address higher-level cognitive reasoning. For instance, even in the high-stakes application of medical diagnosis, models still perform significantly worse than human experts on the PathVQA~\cite{he2020pathvqa} benchmark.

Therefore, future cognitive evaluation must pivot from closed environments to the complex challenges of the real world. \textbf{The benchmarks must transcend basic perception to focus on higher-order cognitive reasoning.} For instance, benchmarks like CA-MER~\cite{han2025benchmarking}, which assesses reasoning in emotionally conflicting scenarios, and CausalVQA~\cite{foss2025causalvqa}, which systematically tests causal understanding, are designed to compel models to move beyond mere ``perception" toward genuine ``reasoning". \textbf{Furthermore, task formats must involve open-ended knowledge integration and creation.} The evaluation emphasis should be on generating logically coherent explanations or executable code, rather than selecting a single correct answer. A forward-looking example is ChartMimic~\cite{yang2024chartmimic}, which requires models to use visual, logical, and programming skills to reproduce a chart.

\section{Conclusion}
\label{sec:Conclusion}
In this survey, we systematically track and summarize the evolution of vision–language interactive reasoning in Multimodal Large Language Models (MLLMs) under a unified From Perception to Cognition framework. To the best of our knowledge, this review offers one of the most comprehensive overviews of how fine-grained Perception underpins robust Cognition. Specifically, we first outline the developmental history of vision–language interaction and present essential background, including core concepts, representative architectures, and evaluation protocols. We then organize the field around two layers: \textbf{On the Perception side}, we synthesize advances in visual encoders, resolution handling, and task-aware vision–language alignment. \textbf{On the Cognition side} , we examine training paradigms for problem decomposition, preference optimization, and dynamic reasoning loops that re-inspect visual evidence. We also catalog benchmarks and applications across documents, charts, scientific reasoning, and healthcare, compile the relevant datasets, and offer fair, comparable analyses where feasible. Finally, we outline concrete future directions: building a truly unified, multi-granular visual encoder, developing latent-space reasoning that plans and searches over internal representations, advancing cross-image relational reasoning and integrating tool-augmented interaction. We also highlight the overarching challenge of closing the perception–cognition gap. This survey aims to serve both newcomers and experienced researchers as a structured, up-to-date reference on current progress and a guide for future work in vision–language interactive reasoning.
\ifCLASSOPTIONcaptionsoff
  \newpage
\fi



{
\bibliographystyle{IEEEtran}
\bibliography{IEEEabrv,ref}
}


\vfill

\end{document}